\RequirePackage{fix-cm}
\documentclass[smallextended]{svjour3}       
\smartqed  
\journalname{Data Mining and Knowledge Discovery}

\usepackage{svg}

\usepackage{amssymb}
\usepackage{amsmath}
\usepackage{dsfont}
\usepackage{graphicx}
\usepackage{caption}
\usepackage{subcaption}
\usepackage{multicol}
\usepackage{comment}

\usepackage[hyphens]{url}
\usepackage[breaklinks]{hyperref}

\usepackage[inline]{enumitem}
\setlist[description]{labelindent=.5em,leftmargin=\dimexpr\parindent+.5em,font=\bfseries}

\usepackage{natbib}
\setcitestyle{authoryear,round,semicolon,aysep={},yysep={, },notesep={, }}
\renewcommand{\cite}{\citep}

\begin{document}

\title{How Robust is your Fair Model? \\
Exploring the Robustness of Prominent Fairness Strategies%
\thanks{%
This research was supported by the ARC Centre of Excellence for Automated Decision-Making and Society, funded by the Australian Government through the Australian Research Council (project number CE200100005). 
During the conception of the work, Wei Shao was with the ARC Centre of Excellence for Automated Decision-Making and Society at RMIT University and was partially supported by the Victorian Government through the Victorian Higher Education Strategic Investment Fund.
Special thanks to Yufan Kang, Damiano Spina and Falk Scholer for helping shape the initial ideas for this paper.%
}}
\titlerunning{How Robust is your Fair Model?}


\author{Edward~A.~Small$^\dagger$\thanks{$^\dagger$~Equal contribution.} \and
Wei~Shao$^\dagger$ \and
Zeliang~Zhang \and
Peihan~Liu \and
Jeffrey~Chan \and 
Kacper~Sokol \and 
Flora~D.~Salim}

\authorrunning{Edward~A.~Small et al.}

\institute{%
Edward~A.~Small (\emph{corresponding author}) \at%
ARC Centre of Excellence for Automated Decision-Making and Society, School of Computing Technologies, RMIT University, Australia\\
\email{edward.small@student.rmit.edu.au}
\and
Wei~Shao \at%
Data61, CSIRO, Australia\\
ARC Centre of Excellence for Automated Decision-Making and Society, School of Computing Technologies, RMIT University, Australia
\and
Zeliang~Zhang \at%
School of Computing Technologies, RMIT University, Australia
\and
Peihan~Liu \at%
Department of Mathematics, University of Michigan, USA
\and
Jeffrey~Chan \at%
ARC Centre of Excellence for Automated Decision-Making and Society, School of Computing Technologies, RMIT University, Australia
\and
Kacper~Sokol \at%
Department of Computer Science, ETH Zurich, Switzerland\\
ARC Centre of Excellence for Automated Decision-Making and Society, School of Computing Technologies, RMIT University, Australia
\and
Flora~D.~Salim \at%
ARC Centre of Excellence for Automated Decision-Making and Society, School of Computer Science and Engineering, UNSW Sydney, Australia
}

\date{Received: date / Accepted: date}

\maketitle

\begin{abstract}
With the introduction of machine learning in high stakes decision-making, ensuring algorithmic fairness has become an increasingly important task. To this end, many mathematical definitions of fairness have been proposed, and a variety of optimisation techniques have been developed, all designed to maximise a given notion of fairness. Fair solutions, however, tend to rely on the quality of training data, and can be highly sensitive to noise. Recent studies have shown that robustness of many such fairness strategies -- i.e., their ability to perform well on unseen data -- is not a given and requires careful consideration. To address this challenge, we propose \textit{robustness ratio}, which is a novel criterion to measure the robustness of diverse fairness optimisation strategies. We support our analysis with multiple extensive experiments on five benchmark fairness data sets, using three prominent fairness strategies, in view of four of the most popular definitions of fairness. Our experiments show that while fairness methods that rely on threshold optimisation (post-processing) mostly outperform other techniques, they are acutely sensitive to noise. This is in contrast to two other methods -- correlation remover (pre-processing) and exponentiated gradient descent (in-processing) -- which become increasingly fairer as the random noise injected into the data becomes larger. Our findings offer a comprehensive overview of fairness strategies that proves invaluable when tasked with choosing the most suitable method for the task at hand. To the best of our knowledge, we are the first to quantitatively evaluate the robustness of fairness optimisation strategies.
\end{abstract}

\section{Introduction}\label{sec1}

As machine learning becomes ubiquitous in high-impact decision-making, enforcing fairness in data-driven predictive models becomes an increasingly important problem to solve. Unfair models can perpetuate discrimination, lead to social injustice, and strengthen unconscious bias by making decisions that are implicitly based on protected attributes.  In fact, we have already seen this happen in healthcare with melanoma detection models performing better on lighter skin tones~\cite{rajkomar2018ensuring}, and in natural language processing with hate speech detection models perpetuating bias related to gendered terms~\cite{NASCIMENTO2022117032}.
Much of recent research regarding fairness aims to develop mathematical definitions of various societal and ethical notions of fairness, as well as build machine learning models that can satisfy these fairness constraints~\cite{bias}. Fairness constraints in existing models can be injected in one of three places: before training, i.e., \textit{pre-processing}; during training, i.e., \textit{in-processing}; or after training, i.e., \textit{post-processing}~\cite{https://doi.org/10.48550/arxiv.2108.04884}.

Although current fairness optimisation strategies are able to mitigate the bias of different learning models, the generalisability of these methods remains an open problem. Many of the mathematical definitions of fairness are group-based, where a target metric is equalised or enforced over sub-populations in the data. We call these sub-populations \textit{protected groups}. Most fairness optimisation strategies heavily rely on these protected class values. Unfortunately, in practice there is always a question as to how reliable this information is; data can be missing, mislabelled, or simply noisy~\cite{eotheory}. These data are often entered by humans, and can contain misleading or false protected information due to the fear of discrimination or disclosure~\cite{SongM}. On top of this, surveyed participants can also give different answers depending on how a question is phrased~\cite{MinsonJuliaA2018Ettt}. This severely limits the generalisability of existing fairness strategies.

The ability of these strategies to generalise to different data qualities can also be referred to as \textit{robustness}. In essence, an algorithm is robust if it:
\begin{enumerate}
    \item performs consistently across different data sets from the same (or similar) input space $\mathcal{X}$ distribution; and
    \item performs consistently for slightly perturbed versions of data sets drawn from the same (or similar) input space $\mathcal{X}$ distribution.
\end{enumerate}
Current fairness strategies are at risk of not remaining fair in view of their definitions in such circumstances, but there is little literature describing how to measure their robustness. 
More specifically, if different strategies have varying degrees of robustness with respect to different learning models or data sets, can we be better informed on which methods are appropriate for a specific problem? Robustness for predictive performance -- which aims to ensure that a model performs consistently across different data sets and noise levels -- and robustness for fairness, while similar in some aspects, are very different in their ultimate aim. In view of this discord, how should we define a robust fairness strategy that balances the two objectives, and measure this phenomenon?

Here, we investigate the performance of different fairness optimisation techniques on benchmark fairness data sets. We propose a new way to inject noise of differing levels into data to measure robustness of fairness, and we explore how this noise injection affects the behaviour of fairness metrics with different models, strategies, and data sets. 

At its core, unfairness or bias between groups is caused by:
\begin{itemize}
    \item prioritising the optimisation of a model for the most dominant group (imbalanced classes);
    \item subtle differences in how each input variable maps to the output space for different groups; and
    \item inconsistencies in cross-correlations or behaviours between groups.
\end{itemize}
When optimising a model for fairness, we adapt the model/data/objective in order to adjust for these discrepancies. However, as we add noise of increasing strength to the underlying data, the random noise becomes more dominant. Therefore, three things occur: %
\begin{enumerate*}[label=(\arabic*)]
    \item 
the signal between the input and output space is weakened;
    \item 
the cross-correlations between variables become less distinct; and
    \item 
the noise becomes the dominant feature in all groups. 
\end{enumerate*}
These factors lead to all groups becoming more similar and, in privacy preserving methods, it becomes impossible to distinguish one group from another. Thus, we argue that the expected behaviour of these models is that they should become fairer as noise levels increase, and that a model that does not follow such a trajectory is not robust with respect to fairness. We go on to justify this motivation mathematically using statistical similarity measures.

In studying this behaviour, we provide a definition of robust fairness optimisation, and find that the bias generated by noise injection is both bounded and related to noise strength. In particular, we provide detailed insights into a popular fairness optimisation strategy, the post-processing method called \textit{threshold optimisation}, under the \textit{equalised odds} fairness constraint to better understand why it is highly sensitive to noise.

The experimental results provide us with three key insights. First, the different fairness constraints can be more easily satisfied for certain data sets, so the measure of fairness can differ depending on how it is defined. Second, the robustness of each of the strategies can differ significantly. Third, post-processing methods are the fairness optimisation strategies most sensitive to noise, often performing the best for small (non-dominant) noise and the worst for large (dominant) noise. 

In summary, the contributions of this paper are as follows:
\begin{enumerate}
    \item we define the expected behaviour of a fair solution that is robust;
    \item we propose a universal metric to measure the robustness of fairness;
    \item we use this metric to empirically compare the robustness of different methods; and
    \item we show that certain methods lack the stability to be robust with respect to fairness.
\end{enumerate}
To the best of our knowledge, this study is the first to analyse the robustness of fairness metrics for different data sets and methods.

\section{Related Work}\label{sec2}

Algorithmic fairness is a popular area of research, especially in recent years. As our understanding of bias and bias mitigation has grown, so too have all of the strategies that can be employed to combat it. As \citet{bias} state, ``bias can exist in many shapes and forms'', and studying these biases is often what motivates a particular solution proposed in fairness literature. 

The definition of fairness is particularly broad, and the definition that is chosen often depends on the modelling problem at hand~\cite{QuyTaiLe2021Asod}. While we acknowledge the vast array of fairness metrics now available -- such as model multiplicity-based fairness~\cite{sokol2024ethical}, preference-based fairness~\cite{https://doi.org/10.48550/arxiv.1707.00010} and individual fairness -- we limit the scope of this study to four of the most common fairness criteria for groups: \textit{demographic parity}, \textit{equalised odds}, \textit{false positive difference}, and \textit{true positive difference}; specifically, we focus on average fairness across the entire group~\cite{group}.

Much like fairness, there are many optimisation methods that can be used to build predictive models. It is therefore impractical to design a study that tests all learning models with all fairness metrics. As such, we explore only binary classification (though this work could be extended to multi-class classification) and select five of the most common machine learning techniques to this end~\cite{methodsurvey}: 
\begin{description}
    \item [LR] Logistic Regression~\cite{alma9921601010301341};
    \item [SVM] Support Vector Machines~\cite{SVM};
    \item [NB] Na\"ive Bayes~\cite{NB};
    \item [SGD] Stochastic Gradient Descent~\cite{SGD}; and
    \item [DT] Decision Tree Classifiers~\cite{DTC}.
\end{description}

While finding methods to define and enforce fairness is being explored for a wide range of problems, the robustness of these solutions currently appears to be mostly an afterthought. Recently, \citet{https://doi.org/10.48550/arxiv.2007.06029} considered a worst-case distribution difference between training and test data, and showed that there is an inherent trade-off between robustness of fairness and accuracy of predictive modelling.

As previously mentioned, we also must consider at what point we should optimise for fairness. This is commonly broken down into three areas: \textit{pre-processing}, \textit{in-processing}, and \textit{post-processing}. 
%
There is a huge variety of processing methods that can be used for pre-processing, in-processing, and post-processing when optimising for fairness. We summarise these methods below and elaborate on them in Section~\ref{sec:fairness_optimisation}.

\paragraph{Pre-processing}
The training data are modified or transformed in such a way as to remove the inherent bias in the data while minimising the loss of any signals that could help a model to learn~\cite{calmon2017optimized,biswas2021fair,farokhi2021optimal}. This allows the model to be trained on data that are less bias, and therefore the model is less likely to exploit the bias to make predictions.

\paragraph{In-processing}
The loss/objective function is altered in such a way that the learning objective balances accuracy and fairness~\cite{kuragano2007curve,kamishima2011fairness,ahn2019fairsight}. In doing this, we allow a model to simultaneously become accurate and fair.

\paragraph{Post-processing}
The predictive model is trained with no knowledge of fairness, and its output is then modified to satisfy a fairness constraint~\cite{kim2019multiaccuracy,lohia2019bias,cui2020xorder}. This is almost equivalent to training a second model that acts as an ``overseer'' of the original model, ensuring that its outputs satisfy the chosen definition of fairness. This is achieved using a combination of thresholding and random allocation.

\section{Preliminaries}\label{sec3}

In this section we introduce the concepts that lay the foundation for the main contributions of this paper. Specifically, we introduce the notation, define the fairness metrics that pertain to the experiments, and explain the fairness processing methods (introduced in the previous section) in more detail.

\subsection{General Notation}
We denote a machine learning model as $f:\mathbb{R}^D\mapsto\mathbb{B}$.
An input data set with $N$ samples and $D$ features is called $\mathbf{X} \subseteq \mathbb{R}^{N\times D}$, where $\mathcal{X} \equiv \mathbb{R}^D$ is the input space, with the labelled output (ground truth) being $\mathbf{Y} \subseteq \mathbb{B}^{N}$, where $\mathcal{Y} \equiv \mathbb{B} \equiv \{0, 1\}$. Inputs $\boldsymbol{x_i}\in \mathbf{X}$ are therefore $\boldsymbol{x_i}\in\mathbb{R}^D$ for $i=\{0,1,\ldots,N\}$ and the corresponding ground truth is $y_i\in\mathbf{Y}$. Since we restrict our experiments to binary classification, $y_i\in\{0, 1\}$. For a matrix, vector, or data set $\mathbf{X}$, we denote the mean of $\mathbf{X}$ as $\mathbf{\bar{X}}$ and the transpose as $\mathbf{X}^T$.

The \textit{protected groups} are the collection of features or attributes that the notion of fairness is optimised with respect to. We call this group $\mathbf{A}$. Some common examples of a protected group are race and sex, as these are prohibited for use as predictive variables in many situations, e.g., employment in view of US law~\cite{fairstandards}.

We denote the probability of an event occurring as $\mathbb{P}\{\cdot\}$, with $\mathbb{P}\{W\vert S,T\}$ meaning ``the probability of $W$ occurring, given that $S$ and $T$ have already occurred''.
Random noise of strength $k$ is denoted as $\epsilon_k$ -- this is defined in more detail in Section~\ref{sec4}. $M(f, \mathbf{X})$ is a general term to describe any given fairness metric with respect to a model $f$ and a set of inputs $\mathbf{X}$. This is elaborated on in the next section.

\subsection{Fairness Metrics\label{sec:fair_met}}

Here, we consider four of the most popular definitions of fairness: %
\emph{demographic parity} (Definition~\ref{def:dp}),
\emph{false positive rate} (Definition~\ref{def:fpr}),
\emph{true positive rate} (Definition~\ref{def:tpr}), and
\emph{equalised odds} (Definition~\ref{def:eo}). With fairness, the objective is to ensure that a probability, or a collection thereof, is equalised between subgroups identified within the protected group. Therefore, we can view this as a minimisation problem. For simplicity, we define $f(\boldsymbol{x})=1$ as a \textit{positive outcome} and $f(\boldsymbol{x})=0$ as a \textit{negative outcome}.
%
In order to refer to different definitions of fairness in a general sense, we specify a \emph{fairness set} (Definition~\ref{def:fairness_set}).
We do this purely for notation purposes as it allows us to refer to all fairness metrics in a general way. 

\begin{definition}[Demographic Parity]\label{def:dp}
\begin{equation}
\begin{aligned}
    M_{dp}(f, \mathbf{X}) = \left\vert\mathbb{P}\{f(\mathbf{X})=1\vert\mathbf{A}=0\} -\mathbb{P}\{f(\mathbf{X})=1\vert\mathbf{A}=1\}\right\vert
    \label{dp}
    \end{aligned}
\end{equation}
For each subgroup in the protected group $\mathbf{A}$, the same proportion of people should receive a positive outcome. Demographic parity is achieved when $M_{dp}(f, \mathbf{X})=0$, and so $\mathbb{P}\{f(\mathbf{X})\vert A\}=\mathbb{P}\{f(\mathbf{X})\}$~\cite{10.1007/978-3-030-93736-2_46}.
\end{definition}

\begin{definition}[False Positive Rate]\label{def:fpr}
\begin{equation*}
\begin{aligned}
    M_{fp}(f, \mathbf{X}) = \left\vert\mathbb{P}\{f(\boldsymbol{X})=1\vert\mathbf{Y}=0,\mathbf{A}=0\}
    - \mathbb{P}\{f(\boldsymbol{X})=1\vert\mathbf{Y}=0,\mathbf{A}=1\}\right\vert
\end{aligned}
\end{equation*}
For each subgroup in the protected group $\mathbf{A}$, the same proportion of people should be incorrectly given a positive outcome. False positive rate is satisfied when $M_{fp}(f, \mathbf{X})=0$~\cite{du2020fairness}.
\end{definition}

\begin{definition}[True Positive Rate]\label{def:tpr}
\begin{equation*}
\begin{aligned}
    M_{tp}(f, \mathbf{X}) = \left\vert\mathbb{P}\{f(\boldsymbol{X})=1\vert\mathbf{Y}=1,\mathbf{A}=0\}
    - \mathbb{P}\{f(\boldsymbol{X})=1\vert\mathbf{Y}=1,\mathbf{A}=1\}\right\vert
\end{aligned}
\end{equation*}
For each subgroup in the protected group $\mathbf{A}$, the same proportion of people should be correctly given a positive outcome. True positive rate is satisfied when $M_{tp}(f, \mathbf{X})=0$~\cite{NEURIPS2021_28267ab8}.
\end{definition}

\begin{definition}[Equalised Odds]\label{def:eo}
\begin{equation}
\begin{aligned}
    &M_{eo}(f, \mathbf{X})\\
    &= \left\vert\mathbb{P}\{f(\boldsymbol{X})=1\vert\mathbf{Y}=y,\mathbf{A}=0\}
    - \mathbb{P}\{f(\boldsymbol{X})=1\vert\mathbf{Y}=y,\mathbf{A}=1\}\right\vert \;\; \forall y
    \label{EO}
\end{aligned}
\end{equation}
For each subgroup in the protected group $\mathbf{A}$, we should have the same proportion of true positive and false positive outcomes. Equalised odds is satisfied when $M_{eo}(f, \mathbf{X})=0$~\cite{eotheory}.
\end{definition}

\begin{definition}[Fairness Set]\label{def:fairness_set}
Based on Definitions~\ref{def:dp}, \ref{def:fpr}, \ref{def:tpr} \& \ref{def:eo}, we define the set of functions $M:f(\mathbf{X})\mapsto [0, 1]$ such that
\begin{equation*}
    M(f,\mathbf{X}) \subseteq  \{M_{dp}, M_{fp}, M_{tp}, M_{eo}\} \text{.}
\end{equation*}
If $M(f, \mathbf{X})=0$, then all fairness definitions are satisfied.
\end{definition}

\subsection{Existing Fairness Optimisation Strategies\label{sec:fairness_optimisation}}

Fairness optimisation strategies are used to enforce the notion of fairness chosen by the stakeholders. There is a variety of processing methods that can be used for optimising for fairness. For this study, we choose one  method from each category based on their popularity. For example, these fairness strategies are used both by IBM -- AI Fairness 360\footnote{\url{https://github.com/Trusted-AI/AIF360}}~\cite{aif360-oct-2018} -- and Microsoft -- Fairlearn\footnote{\url{https://github.com/fairlearn/fairlearn}}~\cite{bird2020fairlearn} -- and are recommended by Google for responsible AI practices: 
\begin{description}
\sloppy
    \item [pre-processing (Definition~\ref{def:preproc})] least-squares fit for correlation removal~\cite{TELLINGHUISEN1994255};
    \item [in-processing (Definition~\ref{def:inproc})] exponentiated gradient~\cite{https://doi.org/10.48550/arxiv.1803.02453}; and
    \item [post-processing (Definition~\ref{def:postproc})] threshold optimisation~\cite{https://doi.org/10.48550/arxiv.1610.02413}.
\end{description}
We compare these strategies to a \emph{baseline} approach (Definition~\ref{def:baseline}).
%
Much like for fairness in Definition~\ref{def:fairness_set}, Definition~\ref{def:function_set} outlines a set of functions so that we can refer to different models in a general way.
Throughout the rest of this section, we offer a deeper explanation of the mechanisms for each approach.

\begin{definition}[Baseline $f_{base}$]\label{def:baseline}
The baseline strategy has no notion of fairness at all. The minimisation or learning algorithm exclusively prioritises predictive performance, and it can therefore exploit the bias in the data if it will assist in making accurate predictions. We therefore use the baseline strategy as a comparison point for all the others strategies. It represents the expected \textit{worst case} unfairness, or the bias that is inherent to the data. We refer to a model using this method as $f_{base}$.
\end{definition}

\begin{definition}[Pre-processing $f_{pre}$]\label{def:preproc}
We approach the correlation removal problem from the least squares perspective. Given a data set $\mathbf{X}$ and protected classes $\mathbf{A}\in\mathbb{R}^{N\times C}$, we remove the protected classes from the data set to get a set $\mathbf{Z}=\mathbf{X} \setminus \mathbf{A}$ that has the protected features removed (so $\mathbf{Z} \subseteq \mathbb{R}^{N\times(D-C)}$). We can then find a data set $\hat{\mathbf{X}}$ that has its dependence on $\mathbf{A}$ (mostly) removed using a linear regression model for each feature against the protected class, i.e.,
\begin{equation*}
    \begin{aligned}
        \mathbf{\hat{X}}_i = \boldsymbol{Z}_i - (\boldsymbol{A}_i-\boldsymbol{\bar{A}})\beta_i \quad 
        \textrm{s.t.} \quad 
        \min_{\beta_i}\vert\vert(\boldsymbol{A}_i-\boldsymbol{\bar{A}})\beta_i - \boldsymbol{Z}_i\vert\vert_2
    \end{aligned}
    \text{~,}
\end{equation*}
where $\mathbf{\bar{A}}$ is the average value of the protected class and $\beta_i\in\mathbb{R}$ for $i=\{1, 2,\ldots, D-C\}$ is the number that satisfies the minimisation constraint for each feature. Fundamentally, we fit a linear model to the centred protected features and calculate the residuals~\cite{TELLINGHUISEN1994255}. The residuals are then removed from the original data, and the model is then trained using either of LR, SVM, NB, SGD, or DT for accuracy only on $\mathbf{\hat{X}}$. We can do this because we trust that the bias has been partially or completely removed using the linear transformation described above. We refer to a model using this method as $f_{pre}$.
\end{definition}

\begin{definition}[In-processing $f_{in}$]\label{def:inproc}
The chosen in-processing method, exponentiated gradient~\cite{KIVINEN19971}, is a type of \textit{mirror descent}~\cite{mirror}. In short, the model is updated by making a series of $K$ predictions and assigning a weight $w^{[i]}_0=\frac{1}{K}$ to each prediction (so $i=[1,2,\ldots,K]$). Weights are updated via the exponential rule by decreasing the value of the weights for a predictors that would have caused increase in the loss function, and vice versa, i.e.,
\begin{equation*}
    w^{[i]}_{t+1}=w^{[i]}_t e^{-\eta\nabla(\mathcal{L}(f))f(X)}
    \text{~,}
\end{equation*}
where $\eta$ is the learning rate and $\sum_{i=1}^K w_t^{[i]} = 1$ and $t$ is the number of iterations.

The crucial thing to understand here is that the fairness constraint is included directly in the objective function $\mathcal{L}$. This means that we optimise for both fairness and accuracy simultaneously, but we prioritise accuracy using a weight constant. This allows the model to slightly violate the fairness rule during learning, provided it does not deviate too far from a fair solution. We refer to a model using this method as $f_{in}$.
\end{definition}

\begin{definition}[Post-processing $f_{post}$]\label{def:postproc}
\sloppy
The post-processing method is threshold optimisation~\cite{https://doi.org/10.48550/arxiv.1610.02413, small2024equalised}. The model is first trained using one of the optimisation methods mentioned in Section~\ref{sec2} (LR, SVM, NB, SGD, or DT) with no notion of fairness -- it is optimised purely for predictive performance (much like the baseline $f_{base}$). After training is complete, the model's output is passed through a second stage. This second stage allows the threshold for positive or negative classification to differ between protected classes in order to fulfil a fairness constraint.

The resulting thresholds mean that the lower bound for a positive outcome is higher for an advantaged class than a disadvantaged class. Conversely, the upper bound for a negative outcome is lower for the disadvantaged class. If a value falls between these bounds, the outcome is random. As a simple example, take $f_s$ to be the first stage of the model and $f_t$ the threshold stage, then $f_s: \mathcal{X} \mapsto [0,1]$ and $f_t: [0,1] \mapsto\{0,1\}$. The post-processing model can establish multiple thresholds and assign a probability distribution that depends on the value of $f_s$ and $\mathbf{A}$. Therefore, if $f_s(x)$ is between thresholds, the output of $f_t$ is random, such that 
\begin{equation}
\begin{aligned}
    \mathbb{P}\{f_t(f_s(x))=1\vert\mathbf{A}=a\} &= p_{1, a}\mathds{1}_{[T_{1, a},1]}(f_s(x)) \\ &+ p_{0, a}\mathds{1}_{[T_{0, a},1]}(f_s(x))
    \end{aligned}
    \label{eq:probs}
\end{equation}
with $T_{y,a}\in[-\infty,\infty]$ and $T_{1,a} \geq T_{0,a}$. Here, $p_{0,a}+p_{1,a}=1$ and
\begin{equation*}
    \mathds{1}_{\mathbf{C}}(x)=
    \begin{cases}
        1 \qquad \textrm{if } x\in C\\
        0 \qquad \textrm{if } x\not\in C
    \end{cases}
\end{equation*}
is the indicator function for a set $\mathbf{C}$. From Equation~\ref{eq:probs}, we can clearly see that 
\begin{equation*}
    \mathbb{P}\{f_t(f_s(x))=1\vert\mathbf{A}=a\} = 
    \begin{cases}
        1 \qquad &\textrm{if } f_s(x) > T_{1,a} \\
        p_{0,a} &\textrm{if } f_s(x)\in[T_{0,a}, T_{1,a}] \\
        0 &\textrm{if } f_s(x) < T_{0,a} \text{~.}
    \end{cases}
\end{equation*}
We refer to a model using this method as $f_{post}$.
For brevity, we refer to $f_t(f_s(x))$ as $\mathbf{F}(x)$.
\end{definition}

\begin{definition}[Function Set]\label{def:function_set}
From Definitions~\ref{def:baseline}, \ref{def:preproc}, \ref{def:inproc} \& \ref{def:postproc}, we define the set of functions $f:\mathcal{X}\mapsto [0, 1]$ where 
\begin{equation*}
    f \subseteq  \{f_{base}, f_{pre}, f_{in}, f_{post}\} \text{~.}
\end{equation*}
\end{definition}

\section{Robustness of Fair Solutions\label{sec4}}

In this study, we move away from monitoring predictive performance and instead examine the model's ability to remain fair under perturbations. We choose to apply random Laplace noise of increasing intensity, as values in the Laplace distribution are collected more heavily around the mean. The noise is injected into the data set, and we measure the changing fairness with respect to the learning model, the fairness constraint, and the location of the constraint.

Robustness $R$ of a learning model $f$ is usually viewed through the lens of its predictive performance, e.g., accuracy or precision~\cite{carlini2017towards}. If a model is robust, its utility remains stable and consistent for small noise injections $\epsilon$, i.e.,
\begin{equation*}
\begin{aligned}
    R(f) &= \frac{1}{N}\sum_{i=1}^N \left\vert f(\mathbf{x}_i) - f(\mathbf{x}_i +\epsilon)\right\vert \\
    &=\frac{1}{N}\left\vert f(\mathbf{X}) - f(\mathbf{X}+\epsilon)\right\vert
    \text{~,}
    \end{aligned}
\end{equation*}
where $\boldsymbol{x}_i \in \mathbf{X} \subseteq \mathbb{R}^{N\times D}$ is an input from a data set. We say that $f$ is robust if $R \leq \delta$ where $\delta$ is some defined tolerance. The robustness of $f$ can depend on the type of noise injection $\epsilon$~\cite{fawzi2016robustness,yu2019interpreting,anderson2020certifying}. Noise can be drawn randomly from a specific distribution~\cite{DIRKTHIELE2007Antd} -- for example, a Gaussian distribution~\cite{lopes2019improving} or a Laplace distribution~\cite{yang2017robust,daxberger2021laplace} -- or it can be crafted in an attempt to fool or destabilise the model -- a strategy often called \textit{adversarial noise}~\cite{carrara2017detecting,kwon2019poster,wang2020targeted}.

Analysing robustness is an important step in safe-guarding a model against adversaries who seek to take advantage of the system, or to ensure continued accuracy across the wider input space. However, this is the very first work that introduces a framework specifically designed to assess the robustness of the fairness measure of a model, rather than its accuracy.

\subsection{Noise\label{sec:sec4:noise}}

Many data-driven models have shown vulnerability to inputs with added noise, such as Gaussian noise, Laplace noise, and Pepper noise~\cite{hendrycks2019benchmarking}. Most studies propose to simply add noise to the input data, such as adding white noise to images~\cite{images} or replacing words with synonyms in natural language processing~\cite{nlp}.
In this paper, we describe noise $\epsilon_k$ of strength $k$ in two slightly different ways depending on the features in $\mathcal{X}$.

\subsubsection{Continuous Data}

\begin{definition}[Continuous Noise]
Define $X_i$ as a data set of continuous variables. We add noise in the following way:
\begin{equation*}
    \mathbf{\tilde{X}}_i=\mathbf{X}_i + \epsilon_k \quad \textrm{with} \quad \epsilon_k\sim L(0,k)
    \text{~,}
\end{equation*}
where $L$ is the Laplace distribution. 
\end{definition}

\begin{figure}[t]
\centering
\includegraphics[width=0.55\textwidth]{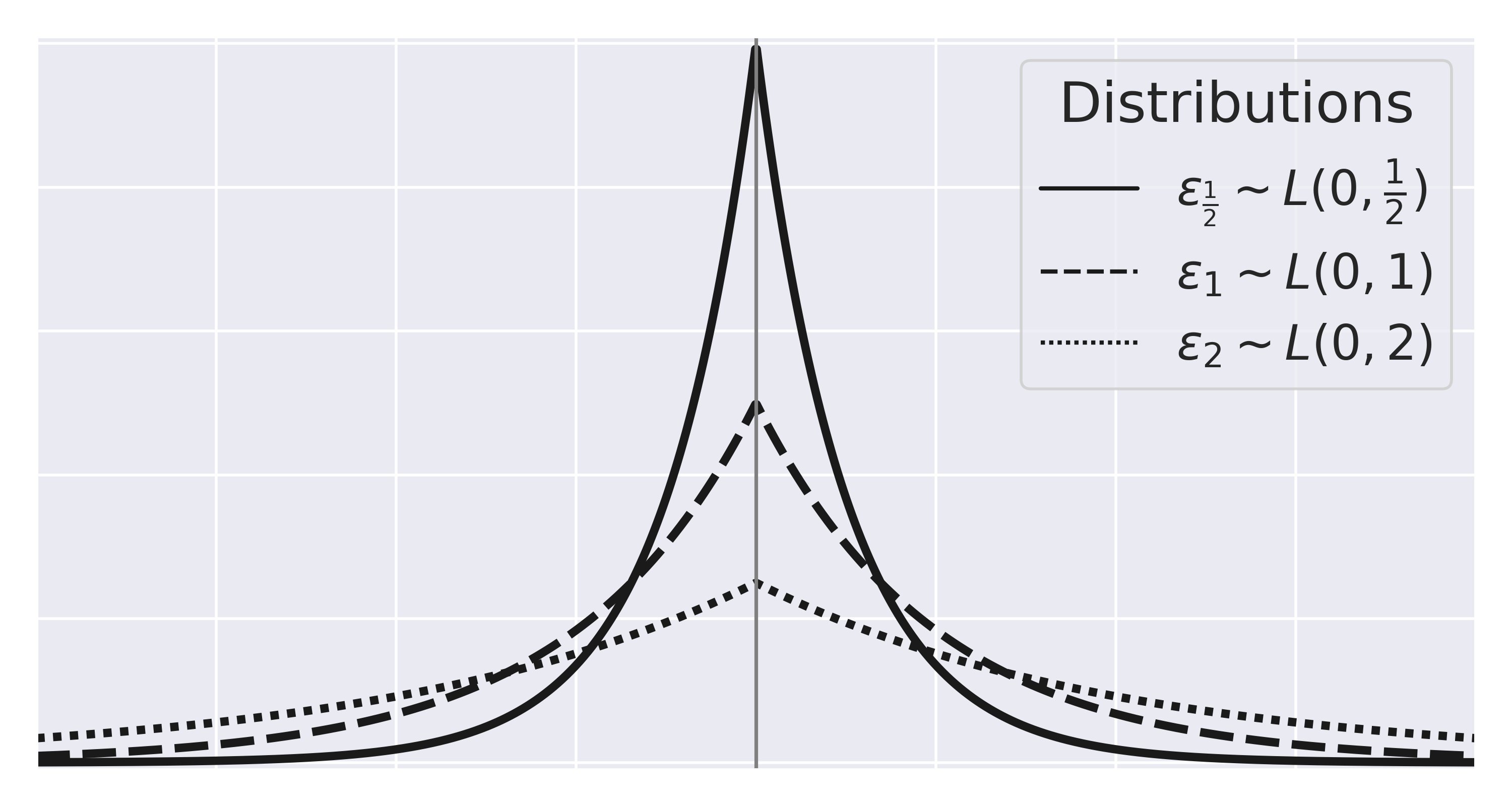}
\caption{Example of the Laplace distribution for different values of $k$. Smaller values of $k$ centre the noise more strongly around $0$.\label{di:lap}}
\end{figure}

As such, the mean of the feature is preserved, but the variance changes. Since the Laplace distribution is defined as
\begin{equation*}
    \epsilon_k(x) = \frac{1}{2k}e^{-\frac{\left\vert x\right\vert }{k}}
\end{equation*}
then the probability of $\epsilon_k$ being in the interval $[a,b]$ is
\begin{equation*}
\begin{aligned}
    \mathbb{P}\{\epsilon_{k}\in[a,b]\}&=\int_a^b\frac{1}{2k}e^{-\frac{\left\vert x\right\vert }{k}} dx\\
    &=\dfrac{b\cdot\left(\frac{1}{2}-\frac{\mathrm{e}^{-\frac{\left\vert b\right\vert }{k}}}{2}\right)}{\left\vert b\right\vert } - \dfrac{a\cdot\left(\frac{1}{2}-\frac{\mathrm{e}^{-\frac{\left\vert a\right\vert }{k}}}{2}\right)}{\left\vert a\right\vert} \text{.}
\end{aligned}
\end{equation*}
See Figure~\ref{di:lap} for the shape of the distribution.

\subsubsection{Discrete Data}

\begin{definition}[Discrete Noise]
    Define a data set $\boldsymbol{x}_i\in\mathbf{X}_i \in \mathbb{R}^N$ and a set $B$ such that
    \begin{equation*}
        \mathbf{X}_i = (\boldsymbol{x}_1, \boldsymbol{x}_2, \ldots, \boldsymbol{x}_N): \boldsymbol{x}_i\in B 
    \end{equation*}
and let $C(b)=\sum_{i=1}^N\mathds{1}_b(\boldsymbol{x}_i)$ be the number of elements in $\mathbf{X}_i$ that are equal to $b\in B$. Given a random variable $s_i\sim B(1, \frac{k}{100})$, we add noise in the following way:
    \begin{equation*}
        \begin{aligned}
            \tilde{\boldsymbol{x}}_i &=  \begin{cases}
            \epsilon_k  \qquad &\textrm{if} \quad s_i = 1 \\ 
            \boldsymbol{x}_i \qquad &\textrm{if} \quad s_i = 0
            \text{~,}
            \end{cases}
        \end{aligned}
    \end{equation*}
where
\begin{equation*}
    \mathbb{P}\{\epsilon_k = b\} = \frac{C(b)}{N} \text{~.}
\end{equation*}
\end{definition}

For an element $\boldsymbol{x}_i$ in a discrete (i.e., categories, objects, or labels) feature column $\mathbf{X}_i \in \mathbb{R}^N$ we use the Bernoulli distribution to create random noise. The distribution is defined via $k$ using:
\begin{equation*}
    \begin{aligned}
        \mathbb{P}(s_i=1)&=\frac{k}{100} \qquad &\textit{for} \quad i=\{1, 2,\ldots, N\}\\
        \mathbb{P}(s_i=0)&=1-\frac{k}{100} \qquad &\textit{for} \quad i=\{1, 2,\ldots, N\} \text{~.}
    \end{aligned}
\end{equation*}
We then sample from the Bernoulli distribution for each element $\boldsymbol{x}_i$ in the discrete feature column $\mathbf{X}_i$, giving
\begin{equation*}
    \boldsymbol{\tilde{x}}_i = 
    \begin{cases}
        \epsilon_k\sim U(\mathbf{X}_i) \qquad &\textrm{if} \quad s_i = 1 \\
        \boldsymbol{x}_i \qquad &\textrm{if} \quad s_i = 0 \text{~.}
    \end{cases}
\end{equation*}

Put more simply, $k$ represents the percentage chance that any element in the discrete set is made to be noisy. The noise $\epsilon_k$ is then sampled from a uniform distribution over the entire feature column $\mathbf{X}_i$. This is to ensure that, even though the data are being randomised, the distribution of the feature does not change (i.e., rare occurrences are still rare, and common occurrences are still common). If there were to be a case where $k \geq 100$, then $\boldsymbol{x}_i = \epsilon_k$ $\forall i$.

\subsection{Expected Behaviour of a Model under Noisy Data}

Before delving into the details of our method, it is important to outline what kind of behaviour is expected of our models given data affected by increasing noise levels to be deemed ``robust w.r.t.\ fairness''. This approach allows us to separate robust models from non-robust models.

As outlined in the introduction, unfairness manifests itself through differences between groups that can be observed in the training data. As such, we can say that, given an input space $\mathcal{X}$ of size $N$ where $\mathcal{X}_i$ for $i=\{1,\ldots,N\}$ are the variables, and we have $M$ distinct protected groups $\mathbf{A}=\{a_1,\ldots,a_M\}$, that
\begin{equation*}
    \mathbb{P}\{\mathcal{X}_i=x\vert A = a_j \} \neq \mathbb{P}\{\mathcal{X}_i=x\vert A = a_k \}
\end{equation*}
for at least one set of values for $i,j,k$. In fact, the way values for each variable are distributed could potentially vary between groups. To quantify this disparity, we can measure the difference between these distributions using the \textit{Bhattacharyya Distance} given in Definition~\ref{def:bah} and built upon the $L^2$ \emph{inner product} outlined in Definition~\ref{def:l2}.

\begin{definition}[$L^2$ Inner Product]\label{def:l2}
    If there exist two real functions $u:\Omega\mapsto\mathbb{R}$ and $v:\Omega\mapsto\mathbb{R}$, then their $L^2$ norm is defined as
    \begin{equation*}
        \langle u,v \rangle_{L^2} = \int_\Omega u(x)v(x) dx \text{~,}
    \end{equation*}
    where $\Omega$ is the domain in which the functions are defined. If $\langle u,v \rangle_{L^2} = 0$, then $u$ and $v$ are orthogonal.
\end{definition}


\begin{definition}[Bhattacharyya Distance] \label{def:bah}
    Given two probability distributions \(p(x)\) and \(q(x)\), we can measure their similarity using the Bhattacharyya distance defined as
    \begin{equation*}
        D_B(p,q)=-\ln{\big(BC(p,q)\big)} \text{~,}
    \end{equation*}
    where
    \begin{equation*}
        BC(p,q)=\int\sqrt{p(x)q(x)}dx
    \end{equation*}
    if $p$ and $q$ are continuous,
    and
    \begin{equation*}
        BC(p,q)=\sum_{x\in\mathbf{X}}\sqrt{p(x)q(x)}
    \end{equation*}
    if $p$ and $q$ are discrete.
Clearly $0 \leq BC \leq 1$, and so $0 \leq D_B \leq \infty$. The closer $D_B$ is to $0$, the more similar the two distributions are.
\end{definition}

\begin{theorem}[Fairness Convergence Under Noise]
\label{thm:noise}
    Take two distributions $p\sim N(\mu_p, \sigma^2_p)$ and $q\sim N(\mu_q, \sigma^2_q)$, which are parameterised by their mean $\mu$ and variance $\sigma$. Adding noise $\delta_k\sim N(0, k^2)$ to both distributions leads to
    \begin{equation*}
        \lim_{k\to\infty} D_B(p+\epsilon_k, q+\epsilon_k) = 0 \text{~.}
    \end{equation*}
(Proof in Appendix~\ref{apx:convergence}.)
\end{theorem}

Notably, $D_B$ tends directly to $0$ and does not oscillate. The convergence of $D_B$ is quadratic in the first term and logarithmic in the second term. The proof is completely intuitive -- as we increase the level of noise, the noise becomes more dominant. Ergo, as noise in the distributions becomes stronger, the two distributions are more alike (albeit the signal becomes weaker).
Even if a feature in $\mathcal{X}$ follows a different distribution due to the value of a protected attribute, adding noise to this feature increases similarity and, therefore, fairness. It follows that we should expect models to display fairer behaviour for increased noise levels.

\subsection{Robustness for Fairness}

We define \textit{robustness for fairness} as the ability of a predictive model to remain fair given perturbed inputs. We measure this as the relative difference between the fairness of a model for $k=0$ and $k>0$, which motivates the definition of \textit{relative robustness} (Definition~\ref{def:relrob}).

\begin{definition}[Relative Robustness]\label{def:relrob}
\begin{equation*}
\begin{aligned}
       R_{k}(f,M,\mathbf{X})&= \mathbb{E}\left[\frac{M(f,\tilde{\mathbf{X}})-M(f,\mathbf{X})}{M(f,\mathbf{X})}\right]+1 \\ 
       &= \int \frac{M(f,\tilde{\mathbf{X}})-M(f,\mathbf{X})}{M(f,\mathbf{X})} d\mathbb{P}\left\{\epsilon_k\right\} + 1
       \text{~,}
       \end{aligned}
\end{equation*}
where $\mathbb{P}(\epsilon_k)$ is the probability distribution of the noise used to create $\tilde{\mathbf{X}}$, as described in Section~\ref{sec:sec4:noise}.
Since $M(f,\tilde{\mathbf{X}}) \in [0,1]$, then
\begin{equation*}
    R_{k}(f,M,\mathbf{X}) \in [0, \infty] \text{~.}
\end{equation*}
\end{definition}

We measure robustness in this particular way for two reasons. One, it preserves the information as to which predictive model gets less fair and which model becomes more fair as $k$ changes. Two, it allows us to specifically measure how much worse a metric is compared to its fairness at $k=0$. The expected behaviour is:
\begin{enumerate}
    \item If $R_{k}(f,M,\mathbf{X})\to 1$ as $k\to\infty$, the method remains stable, i.e., it does not improve or worsen with increasing noise.
    \item If $R_{k}(f,M,\mathbf{X})\to 0$ as $k\to\infty$, the method becomes perfectly fair with increasing noise.
    \item If $R_{k}(f,M,\mathbf{X})\to a$ with $1>a>0$ as $k\to\infty$, the method performs fairer than $k=0$, with a factor of ${a}\cdot M(f,\mathbf{X})$ being the best case scenario for performance. In essence, the method performs $\frac{1}{a}$ times better for strong noise.
    \item If $R_{k}(f,M,\mathbf{X})\to b$ with $b>1$ as $k\to\infty$, the method performs less fair as noise increases, with a factor of $b\cdot M(f,\mathbf{X})$ being the worst case scenario for performance.
\end{enumerate}

\begin{definition}[Robustness Ratio]
If $\boldsymbol{x}_i$ is a sample from a data set $\mathbf{X} \subseteq \mathbb{R}^{N\times D}$, we can approximate $R_k(f, M, \mathbf{X})$ using the discrete formulation
\begin{equation}
    R_k^{[K]}(f,M,\mathbf{X}) =\sum_{j=1}^K\sum_{i=1}^N\frac{M(f,\tilde{\boldsymbol{x}}_{i,j})}{M(f,\boldsymbol{x_i})} \text{~.}
    \label{dis_robustness_eq2}
\end{equation}
(Proof in Appendix~\ref{apx:ratio}.)
\end{definition}

\begin{figure}[t]
\centering
  \includegraphics[width=0.9\textwidth]{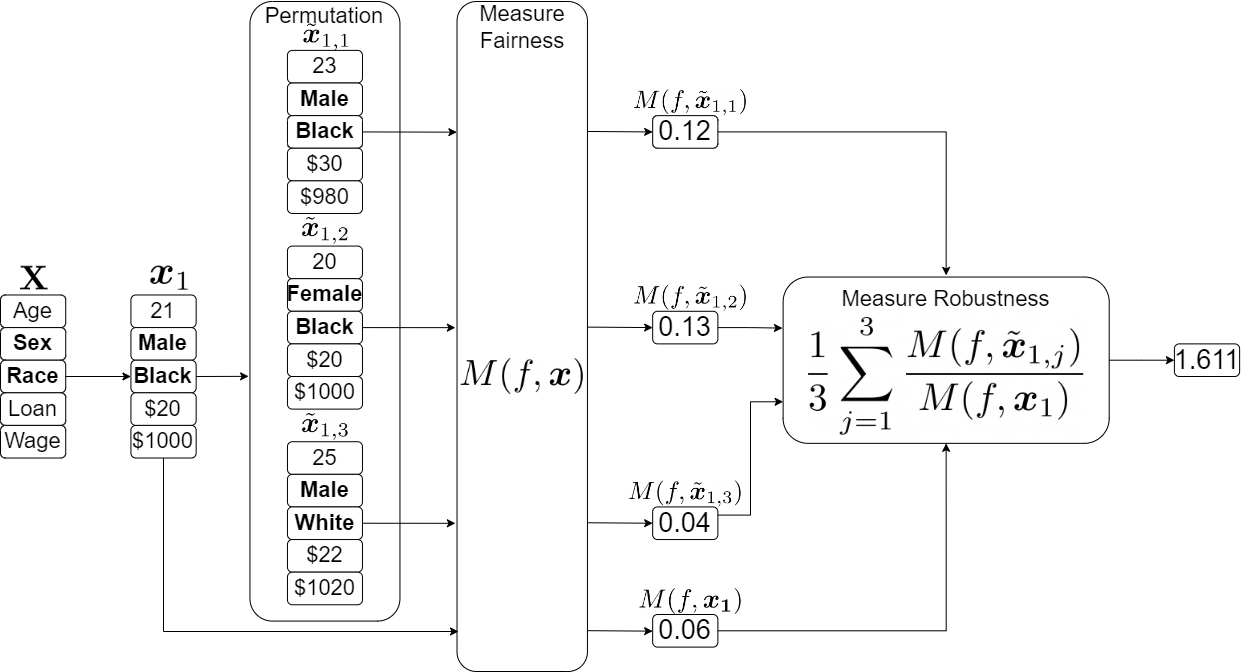}
  \caption{Pipeline example for measuring the robustness of a model $f$ for a single input $\boldsymbol{x}_1$ with $K=3$. If we only have one sample, i.e., $N=1$, so the first sum over the data set can be dropped from the robustness ratio. The input is perturbed slightly, and the function $M$ measures the fairness for both the original input and the perturbed inputs $\tilde{\boldsymbol{x}}_{1,i}$. The perturbed inputs are then compared to the original input. Protected features are labelled in \textbf{bold}.}
  \label{pipeline}
\end{figure}

We calculate this $K$ times for all inputs $\boldsymbol{x_i}\in\mathbf{X} \subseteq \mathbb{R}^{N\times D}$, and take an average to approximate the robustness. The pipeline is shown succinctly in Figure~\ref{pipeline}.

\section{Experiments}

The aim of the experiments is to explore how models that are optimised for a particular notion of fairness behave, with respect to their fairness definition, as input data become increasingly noisy. Specifically, we search for models where the solution becomes \textbf{no less fair} as data become noisier. We call these models robust fair models. 

To measure the robustness of the fairness methods, we used the robustness ratio (as defined in Equation~\ref{dis_robustness_eq2}) with varying noise levels across multiple data sets, fairness metrics, and learning models. We vary  the value of $K$ in reflection of compute time (which depends on the data sample size, the learning model, and the fairness metric), but $k\in[0,10]$ remains consistent. In total, we leverage:
\begin{itemize}
    \item five real-world data sets;
    \item four fairness metrics (Section~\ref{sec:fair_met});
    \item five learning models (Section~\ref{sec2}); and 
    \item four fairness optimisation strategies (Section~\ref{sec:fairness_optimisation}).
\end{itemize}

For the sake of brevity, we do not include all the results in the main body of this paper. Instead, we limit our discussions to show one example of each of the five learning algorithms (LR, SVM, NB, SGD, and DT) over five of the most common fairness data sets (Adult Income, COMPAS, Dutch, Law School, and Bank Marketing) using all four of the predefined notions of fairness from Section~\ref{sec:fair_met} and all four of the strategies outlined in Section~\ref{sec:fairness_optimisation}. Ultimately, we investigate 80 results for fairness and the corresponding 80 results for robustness, with the full body of results available online. 
To this end, we have created a robustness package called \textit{FairR} -- available on GitHub\footnote{\url{https://github.com/TeddyZander/FairR}} -- which utilises the \textit{Fairlearn}~\cite{bird2020fairlearn} package\footnote{\url{https://github.com/fairlearn/fairlearn}}. 

\subsection{Data Sets}

Many of the data sets come from the US Census, which are specifically kept and maintained for fairness benchmark testing. On top of this, we also run experiments on other benchmark data sets that are widely used in the fair machine learning research.
For the American Community Service (ACS) style data sets, we used the census information from the state of California, as it is very dense and is recommended by the authors\footnote{\url{https://github.com/zykls/folktables}}.

\paragraph{Adult Income}
Known as \textit{ACS Income}, this is a modern version of the \textit{UCI Income} data set~\cite{kohavi1996scaling}. The goal with this data set is to create a model that predicts whether an individual's income exceeds \$50,000 a year using 378,817 samples and ten features (eight discrete, two continuous). The protected feature is sex.
This data set is an improvement over the previous version in several ways. First, it is significantly denser, with the full data set containing over a million samples compared to the old data set's 49,531. Second, while disparities between protected classes still exist, they are weaker in the newer data~\cite{https://doi.org/10.48550/arxiv.2108.04884}.

\paragraph{Bank Marketing}
The \textit{Bank Marketing} data set\footnote{\url{https://archive.ics.uci.edu/ml/datasets/bank+marketing}} was used by a Portuguese bank to create a model that can predict whether or not an individual will subscribe to a long term deposit. The data set has 45,211 samples with 17 features (ten continuous, seven discrete). The protected feature is marital status, and as such we filter the data beforehand to remove any samples where marriage status is unknown.

\paragraph{COMPAS}
The Correctional Offender Management Profiling for Alternative Sanctions (\textit{COMPAS}) data set\footnote{\url{https://github.com/propublica/compas-analysis}} was used to train a real predictive model that assisted judges and parole officers. The goal is to predict whether or not a criminal will re-offend within the next two years (two year recidivism rate). The scenario was a landmark case in unfairness when it was shown that the rate of false positives for black men was twice as high as it was for white men~\cite{compas}.
We use the full 7,214 samples; however, the raw data contain plenty of features that are not predictive, and so we trim down the raw data set to nine features (eight discrete, one continuous), and we use race as the protected feature. We also pre-process the data to make race a binary attribute: white and non-white.

\paragraph{German Credit}
Also known as the \textit{StatLog} data set; the objective is to assess whether or not an individual represents a \emph{good} or \emph{bad} credit risk. The data set contains 1,000 samples with 20 features (13 discrete, seven continuous). The protected attribute is a combination between marital status and sex, although an argument could be made that the binary \emph{foreign worker} feature could also be a protected feature.
The class imbalance in this data set is very severe, and so we equalise the classes prior to model training. We also filter the data such that the protected feature is sex only, rather than the combined data of sex and marital status.

\paragraph{Law School}
The Law School Admission Council (\textit{LSAC}) data set was created in 1998 to respond to ``rumors and anecdotal reports suggesting bar passage rates were so low among examinees of color that potential applicants were questioning the wisdom of investing the time and resources necessary to obtain a legal education''\footnote{\url{https://eric.ed.gov/?id=ED469370}}. The goal of the model is to predict whether or not an individual would pass the bar exam. The data include 20,798 samples, with 12 features (nine discrete, three continuous), with the protected class being \emph{race}.

\subsection{Results}
\subsubsection{Evaluation on Adult Income}

The results for ACS Income are restricted to SGD, with $K=100$. We choose SGD for this data set because it is by far the largest and, as such, the compute time decreased more significantly (in absolute terms) when using SGD here than with any other data set.

\begin{figure}
         \begin{multicols}{2}
     \centering
     \begin{subfigure}[b]{0.475\textwidth}
         \centering
         \includegraphics[width=.875\textwidth]{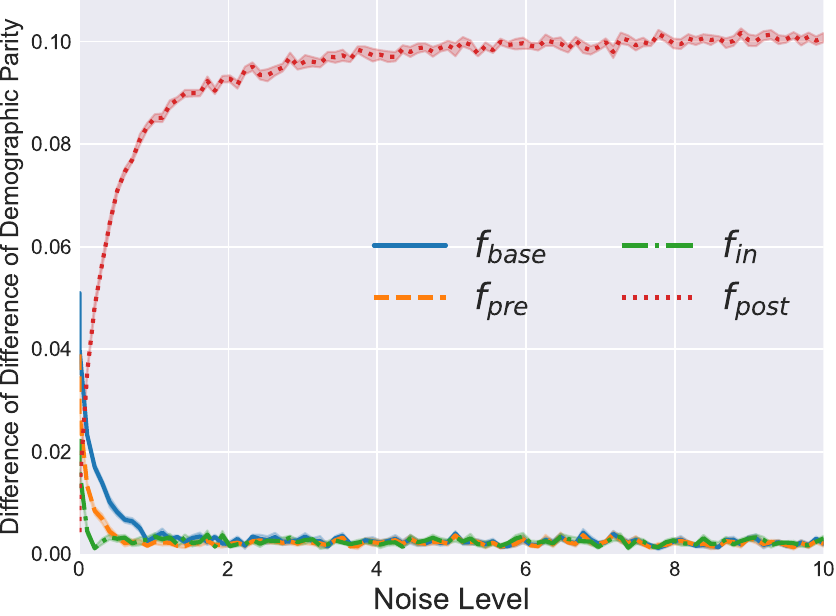}
         \caption{Demographic Parity fairness.}
         \label{adult_fair_dp}
     \end{subfigure}
     \begin{subfigure}[b]{0.475\textwidth}
         \centering
         \includegraphics[width=.875\textwidth]{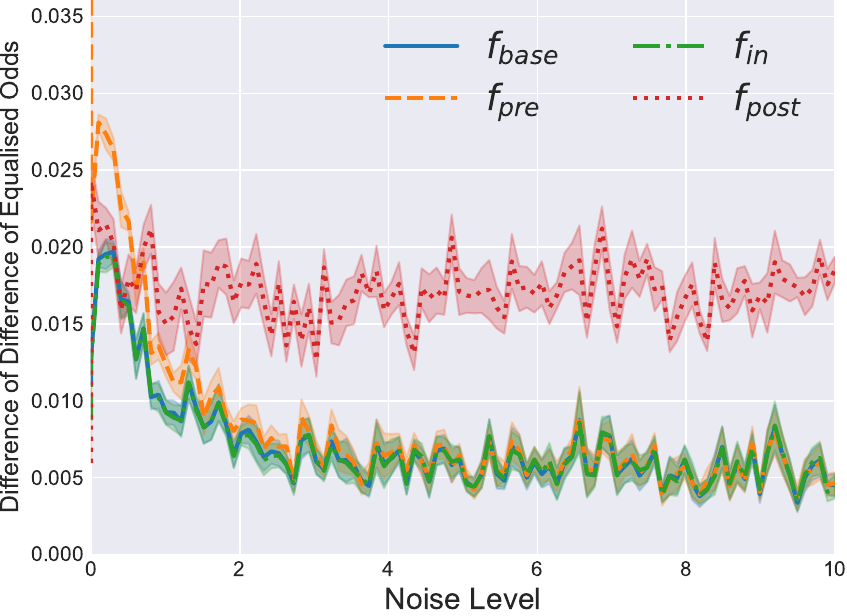}
         \caption{Equalised Odds fairness.}
         \label{adult_fair_eo}
     \end{subfigure}
     \begin{subfigure}[b]{0.475\textwidth}
         \centering
         \includegraphics[width=.875\textwidth]{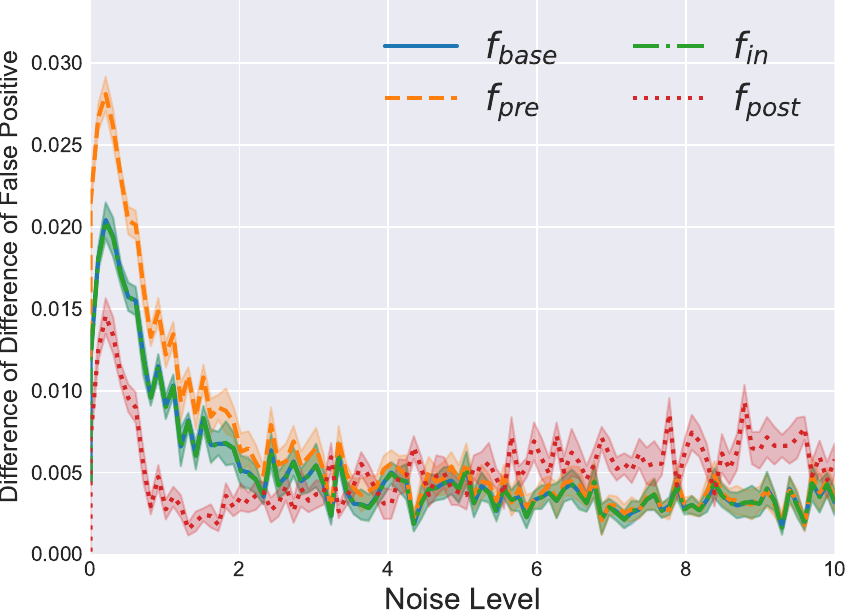}
         \caption{False Positive fairness.}
         \label{adult_fair_fp}
     \end{subfigure}
     \begin{subfigure}[b]{0.475\textwidth}
         \centering
         \includegraphics[width=.875\textwidth]{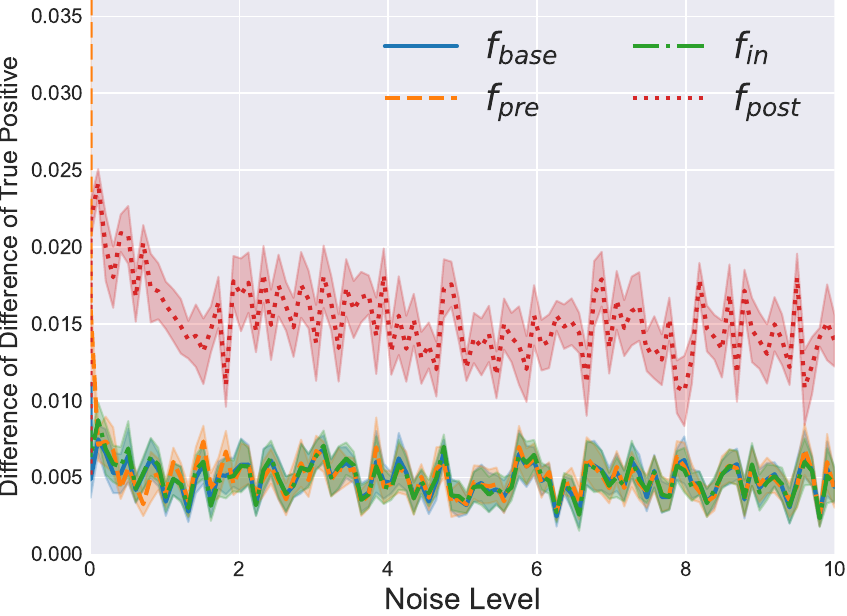}
         \caption{True Positive fairness.}
         \label{adult_fair_tp}
     \end{subfigure}
     \begin{subfigure}[b]{0.475\textwidth}
         \centering
         \includegraphics[width=.875\textwidth]{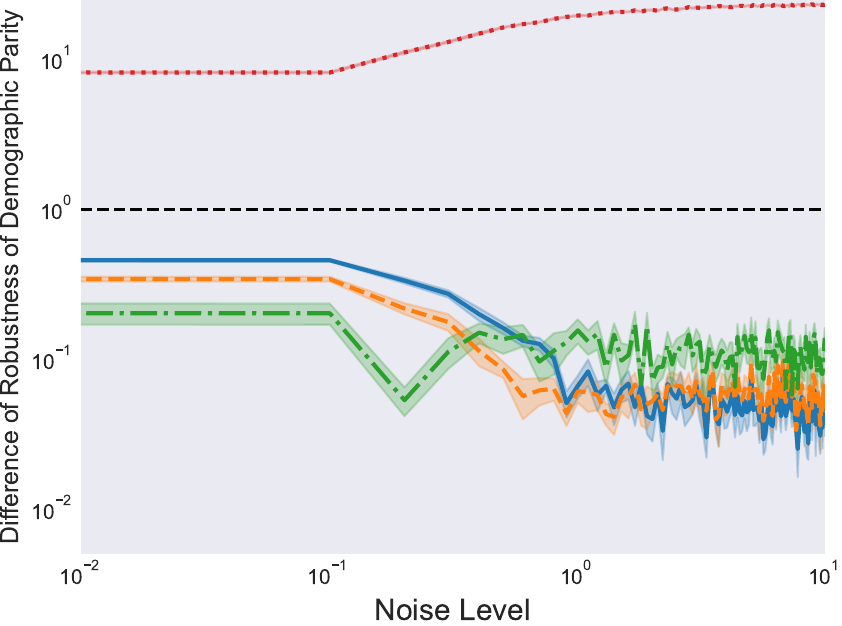}
         \caption{Demographic Parity robustness.}
         \label{adult_rob_dp}
     \end{subfigure}
     \begin{subfigure}[b]{0.475\textwidth}
         \centering
         \includegraphics[width=.875\textwidth]{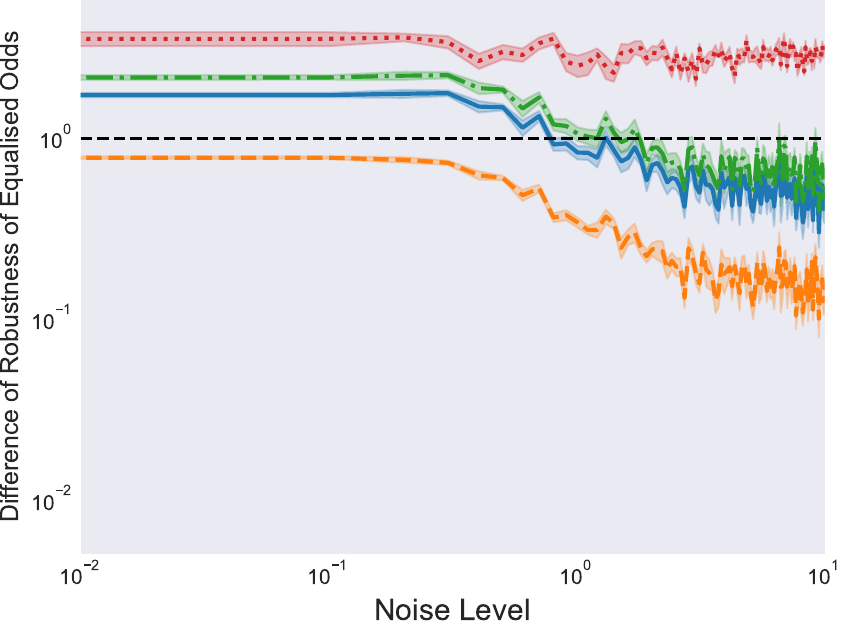}
         \caption{Equalised Odds robustness.}
         \label{adult_rob_eo}
     \end{subfigure}
     \begin{subfigure}[b]{0.475\textwidth}
         \centering
         \includegraphics[width=.875\textwidth]{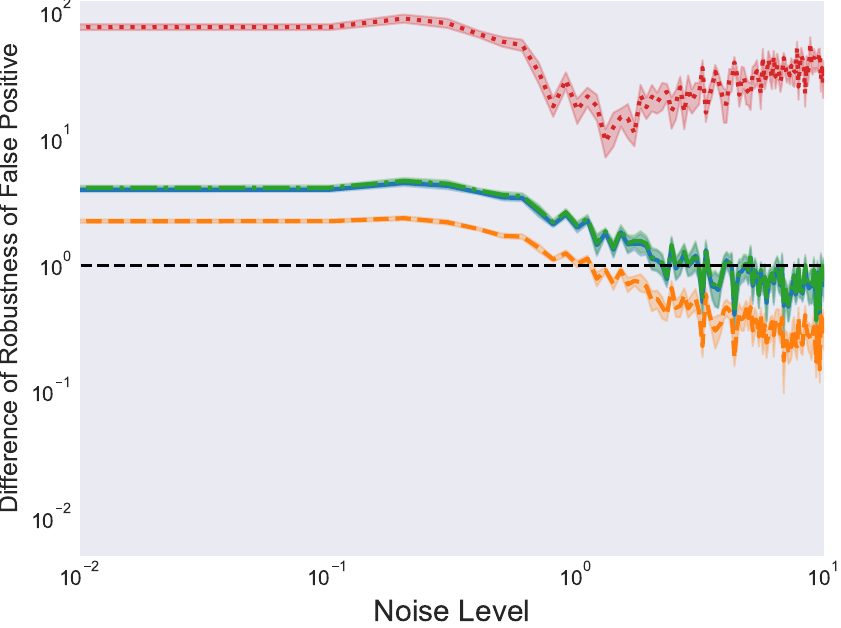}
         \caption{False Positive robustness.}
         \label{adult_rob_fp}
     \end{subfigure}
     \begin{subfigure}[b]{0.475\textwidth}
         \centering
         \includegraphics[width=.875\textwidth]{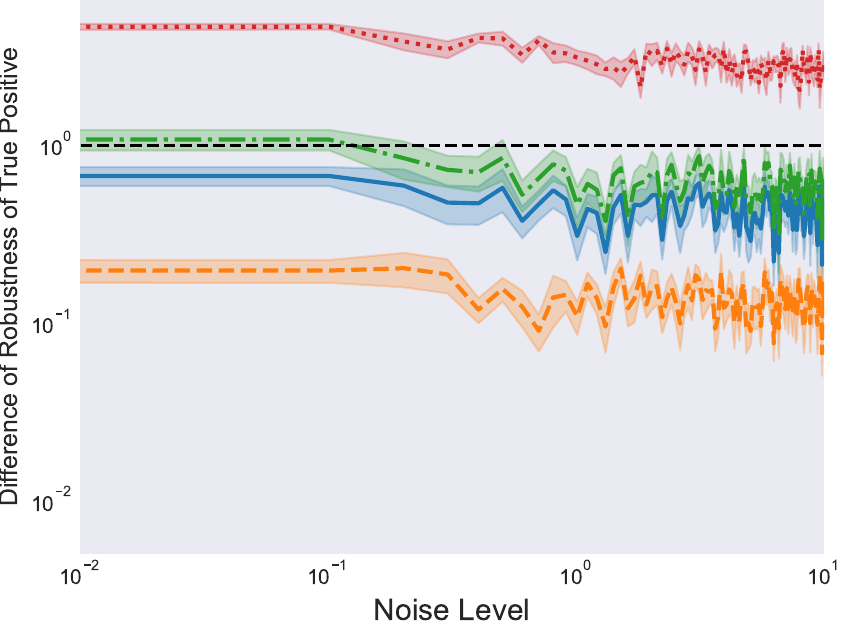}
         \caption{True Positive robustness.}
         \label{adult_rob_tp}
     \end{subfigure}
         \end{multicols}
          \caption{Fairness and robustness of fairness for various strategies using the Adult Income data set with stochastic gradient descent.}
     \label{adult}
\end{figure}

\paragraph{Fairness -- Adult Income}
The evaluation of the fairness for Adult Income data set is shown in Figure~\ref{adult}(\subref{adult_fair_dp}, \subref{adult_fair_eo}, \subref{adult_fair_fp} \& \subref{adult_fair_tp}). The first observation is that $f_{pre}$ does not always perform better than the baseline method for $k=0$. More specifically, it only performs better for demographic parity. All the methods perform well on an absolute scale for true positive. $f_{post}$ clearly shows a different behaviour to the other methods.

\paragraph{Robustness -- Adult Income}
The behaviour from $f_{post}$ is more clear when we measure the robustness ratio shown in Figure~\ref{adult}(\subref{adult_rob_dp}, \subref{adult_rob_eo}, \subref{adult_rob_fp} \& \subref{adult_rob_tp}). While it performs well for $k=0$ for all fairness metrics, it is the only method for all four metrics that consistently performs more poorly as noise increases. The worst absolute score is in demographic parity,  as is the biggest relative change. As $k\to10$, $R_k(f_{post}, M_{dp}, \mathbf{X})\to27$, and even for low noise $R_{1}(f_{post}, M_{dp}, \mathbf{X})\approx100$, which is an unusually small amount of noise for such a large change in performance.
All other methods show reasonably stable performance for large noise as $R_{10}(f, M, \mathbf{X})<1$ for all metrics. The false positive method seems to show the worst stability for small levels of noise as all methods exceed $1$ for $k\in(0, 1)$.

\subsubsection{Evaluation on COMPAS}

The results for the COMPAS data set are restricted to DT, with $K=1000$. We choose DT for this data set because the data set contains a large amount of discrete information that has multiple categories. We can also choose a large number for $K$ because the sample size is reasonably small, and optimisation for DT is fast.

\begin{figure}
         \begin{multicols}{2}
     \centering
     \begin{subfigure}[b]{0.475\textwidth}
         \centering
         \includegraphics[width=.875\textwidth]{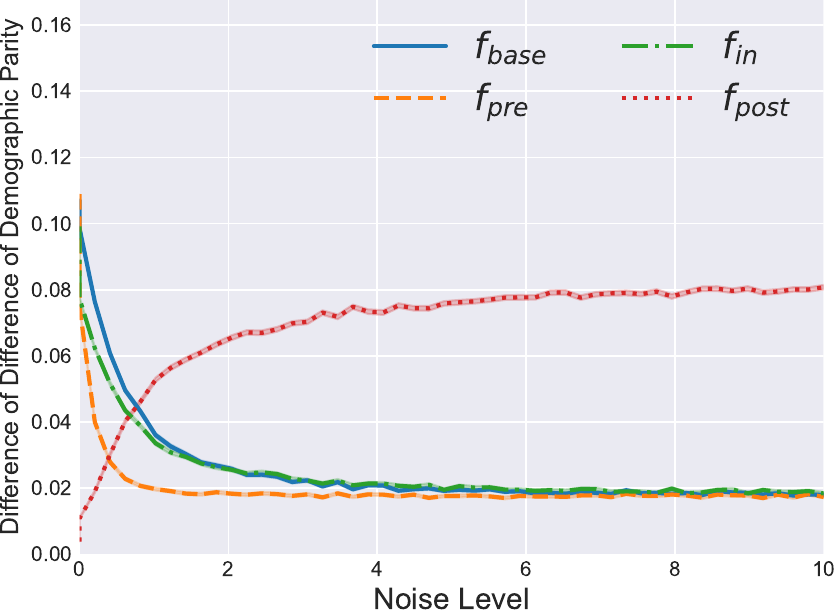}
         \caption{Demographic Parity fairness.}
         \label{compas_fair_dp}
     \end{subfigure}
     \begin{subfigure}[b]{0.475\textwidth}
         \centering
         \includegraphics[width=.875\textwidth]{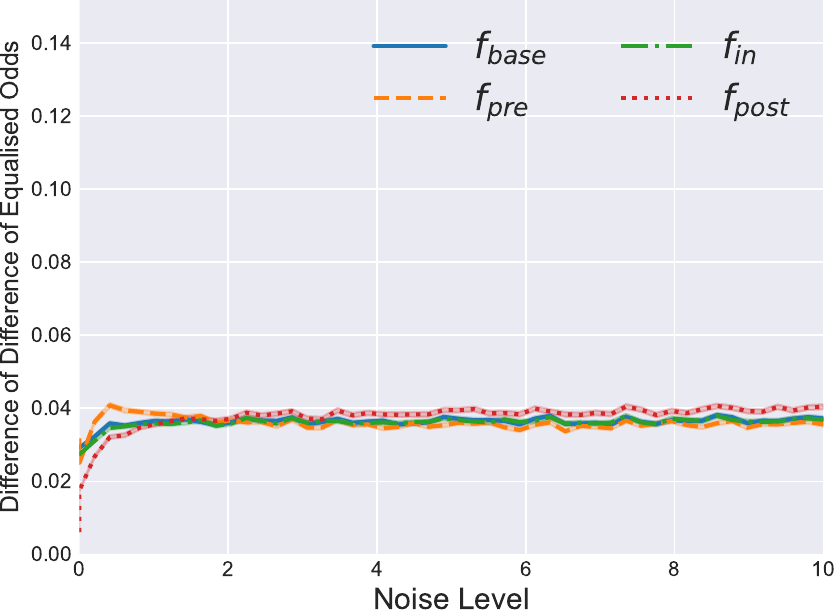}
         \caption{Equalised Odds fairness.}
         \label{compas_fair_eo}
     \end{subfigure}
     \begin{subfigure}[b]{0.475\textwidth}
         \centering
         \includegraphics[width=.875\textwidth]{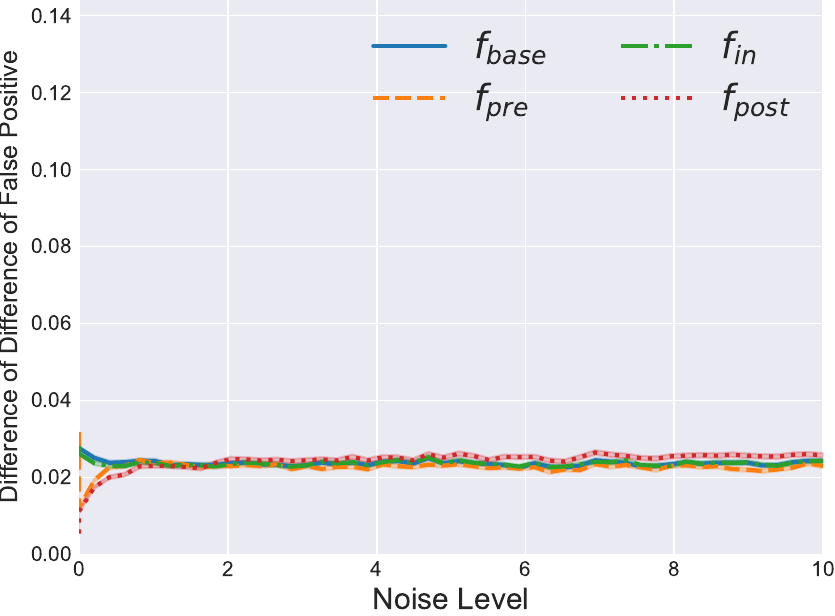}
         \caption{False Positive fairness.}
         \label{compas_fair_fp}
     \end{subfigure}
     \begin{subfigure}[b]{0.475\textwidth}
         \centering
         \includegraphics[width=.875\textwidth]{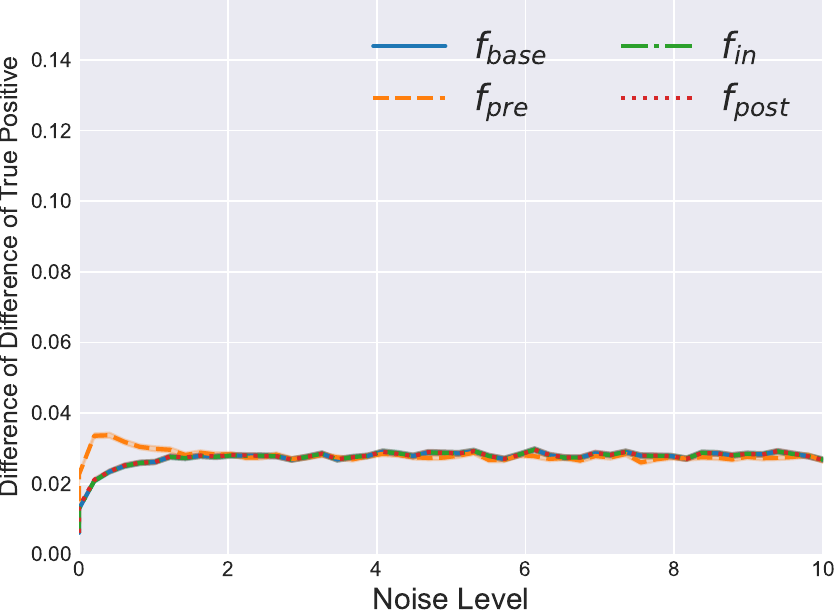}
         \caption{True Positive fairness.}
         \label{compas_fair_tp}
     \end{subfigure}
     \begin{subfigure}[b]{0.475\textwidth}
         \centering
         \includegraphics[width=.875\textwidth]{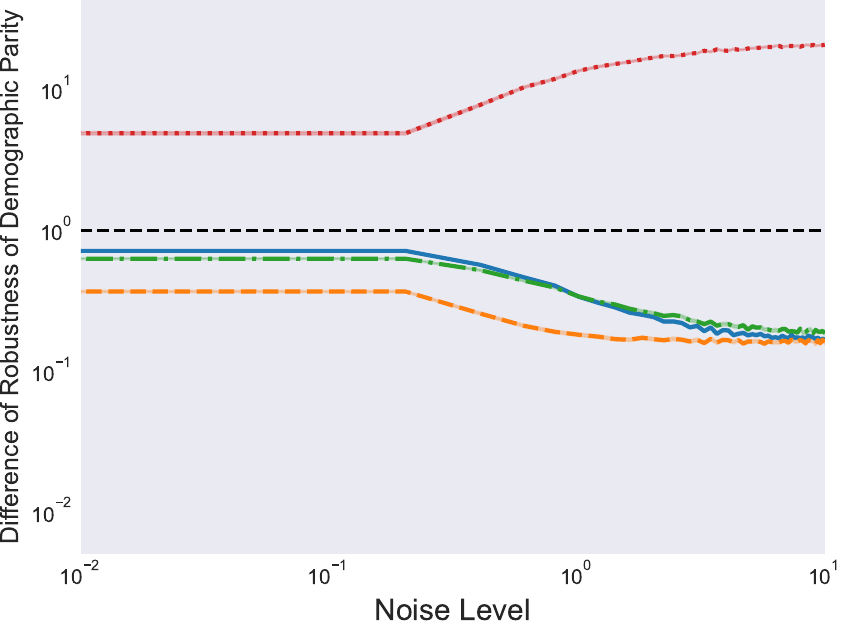}
         \caption{Demographic Parity robustness.}
         \label{compas_rob_dp}
     \end{subfigure}
     \begin{subfigure}[b]{0.475\textwidth}
         \centering
         \includegraphics[width=.875\textwidth]{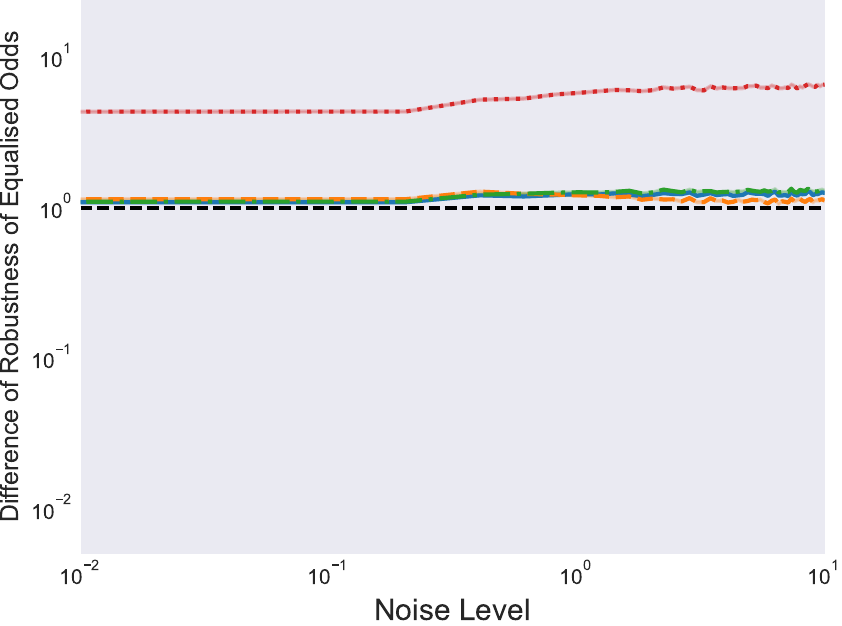}
         \caption{Equalised Odds robustness.}
         \label{compas_rob_eo}
     \end{subfigure}
     \begin{subfigure}[b]{0.475\textwidth}
         \centering
         \includegraphics[width=.875\textwidth]{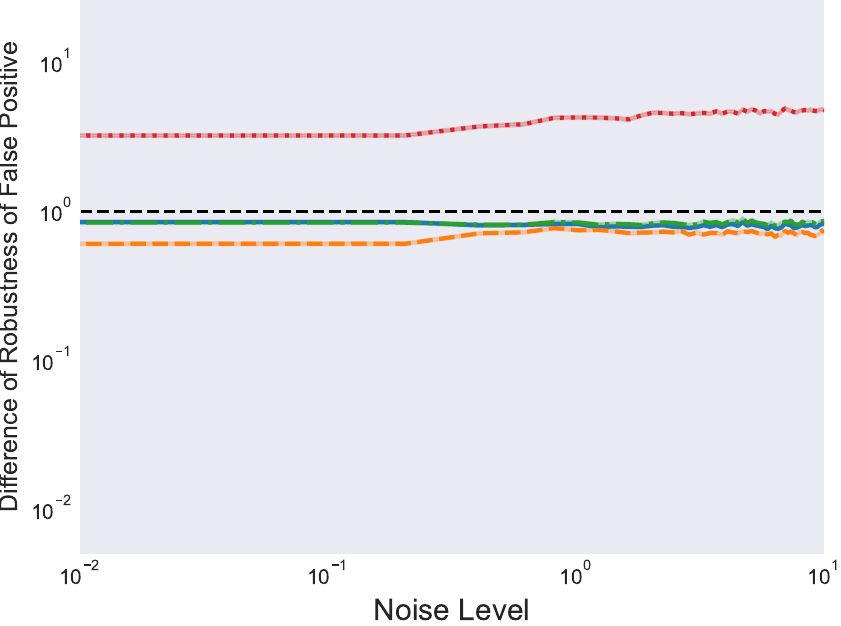}
         \caption{False Positive robustness.}
         \label{compas_rob_fp}
     \end{subfigure}
     \begin{subfigure}[b]{0.475\textwidth}
         \centering
         \includegraphics[width=.875\textwidth]{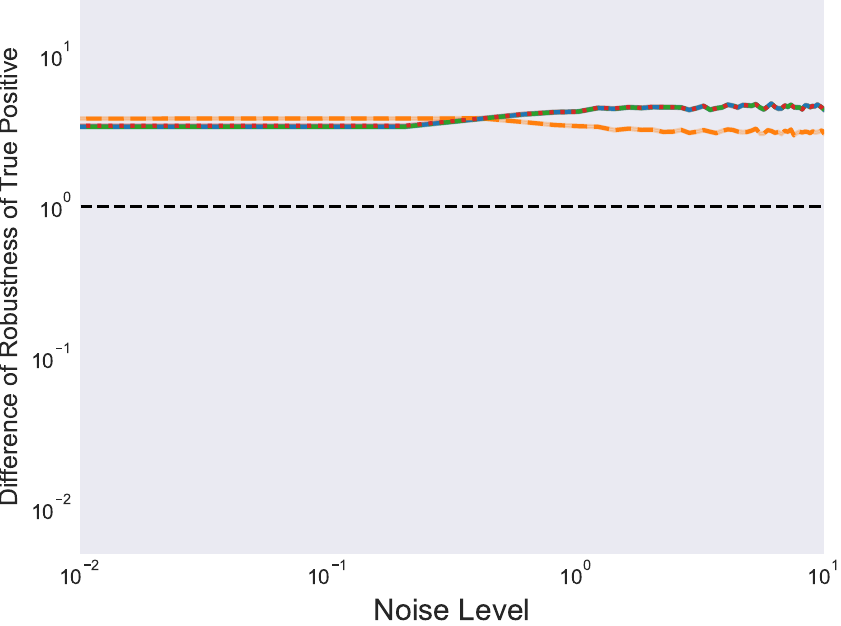}
         \caption{True Positive robustness.}
         \label{compas_rob_tp}
     \end{subfigure}
         \end{multicols}
          \caption{Fairness and robustness of fairness for various strategies using the COMPAS data set with decision tree classification.}
     \label{compas}
\end{figure}

\paragraph{Fairness -- COMPAS}

The evaluation for the fairness of the COMPAS data set is found in Figure~\ref{compas}(\subref{compas_fair_dp}, \subref{compas_fair_eo}, \subref{compas_fair_fp} \& \subref{compas_fair_tp}). The demographic parity pattern is very similar to the Adult Income data set in that:
\begin{itemize}
    \item $M_{dp}(f_{post}, \mathbf{X})$ performs the best at $k=0$;
    \item $M_{dp}(f_{pre}, \mathbf{X})$ and $M_{dp}(f_{in}, \mathbf{X})$ perform very similarly to $M_{dp}(f_{base}, \mathbf{X})$ at $k=0$; and
    \item As $k\to10$, $M_{dp}(f_{post}, \mathbf{X})$ performs worse than all the other methods.
\end{itemize}
$M_{dp}(f_{post}, \mathbf{X})$ has the lowest unfairness at $k=0$, but, for all metrics other than demographic parity, the performance of all strategies is similar for large noise.

\paragraph{Robustness -- COMPAS}

The breakdown of the robustness of the COMPAS data is shown in Figure~\ref{compas}(\subref{compas_rob_dp}, \subref{compas_rob_eo}, \subref{compas_rob_fp} \& \subref{compas_rob_tp}). The first observation is that the performance degrades roughly by a factor of 20 for $f_{post}$ in demographic parity.
We can also see that, while $f_{post}$ has the largest relative change from $k=0$ for equalised odds, they all follow a very similar pattern, just on different scales. Finally, we can also see that all methods perform worse as $k$ increases for the true positive fairness metric.

\subsubsection{Bank Marketing}

The results for the Bank Marketing data set are restricted to NB, with K = 100. We choose NB for this data set because the data set contains plenty of continuous information.

\begin{figure}
         \begin{multicols}{2}
     \centering
     \begin{subfigure}[b]{0.475\textwidth}
         \centering
         \includegraphics[width=.875\textwidth]{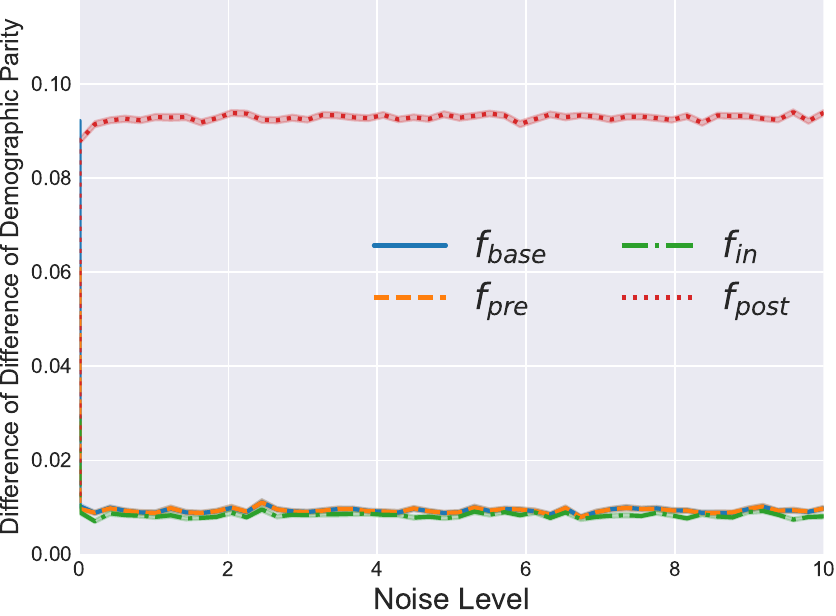}
         \caption{Demographic Parity fairness.}
         \label{bank_fair_dp}
     \end{subfigure}
     \begin{subfigure}[b]{0.475\textwidth}
         \centering
         \includegraphics[width=.875\textwidth]{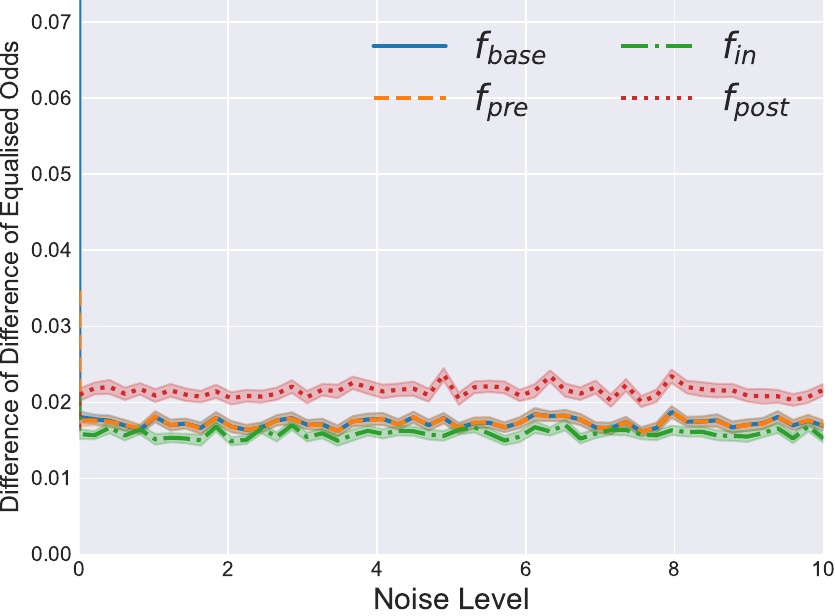}
         \caption{Equalised Odds fairness.}
         \label{bank_fair_eo}
     \end{subfigure}
     \begin{subfigure}[b]{0.475\textwidth}
         \centering
         \includegraphics[width=.875\textwidth]{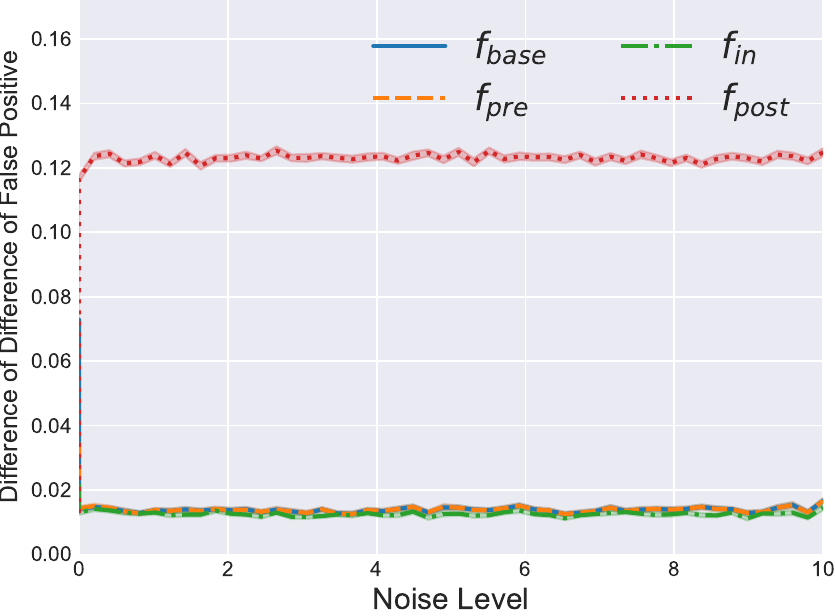}
         \caption{False Positive fairness.}
         \label{bank_fair_fp}
     \end{subfigure}
     \begin{subfigure}[b]{0.475\textwidth}
         \centering
         \includegraphics[width=.875\textwidth]{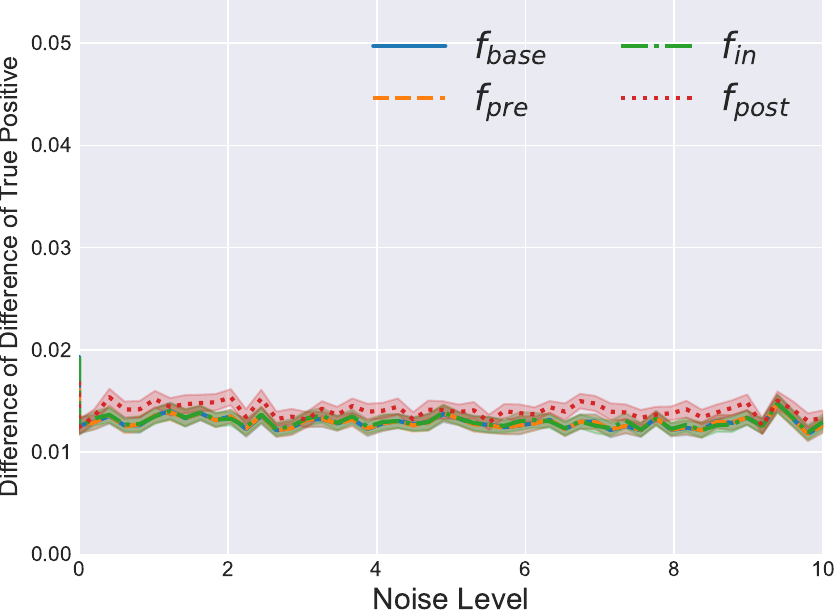}
         \caption{True Positive fairness.}
         \label{bank_fair_tp}
     \end{subfigure}
     \begin{subfigure}[b]{0.475\textwidth}
         \centering
         \includegraphics[width=.875\textwidth]{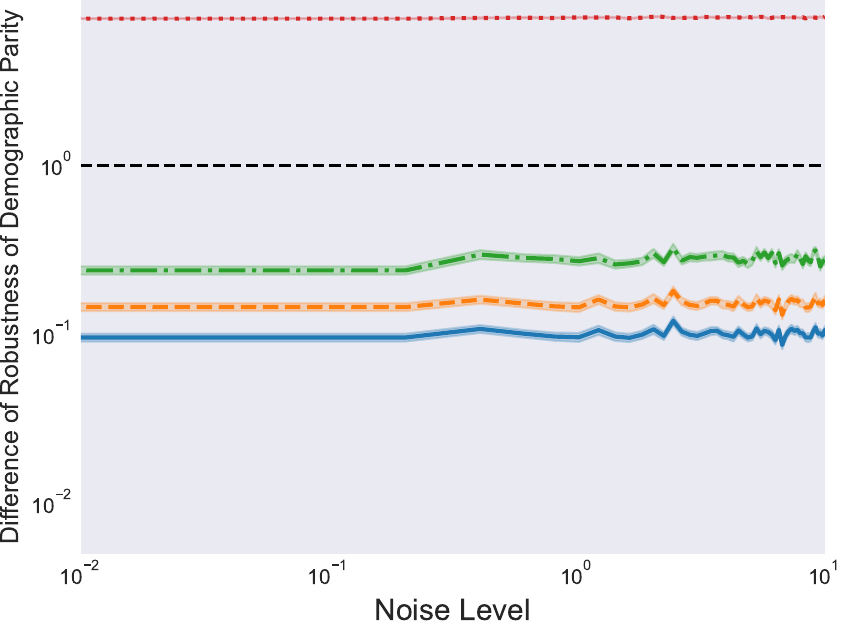}
         \caption{Demographic Parity robustness.}
         \label{bank_rob_dp}
     \end{subfigure}
     \begin{subfigure}[b]{0.475\textwidth}
         \centering
         \includegraphics[width=.875\textwidth]{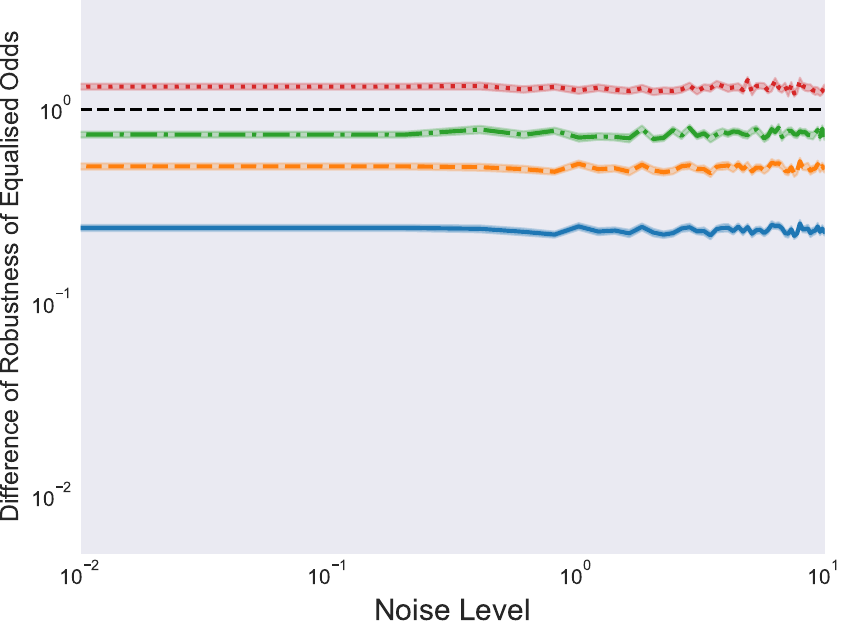}
         \caption{Equalised Odds robustness.}
         \label{bank_rob_eo}
     \end{subfigure}
     \begin{subfigure}[b]{0.475\textwidth}
         \centering
         \includegraphics[width=.875\textwidth]{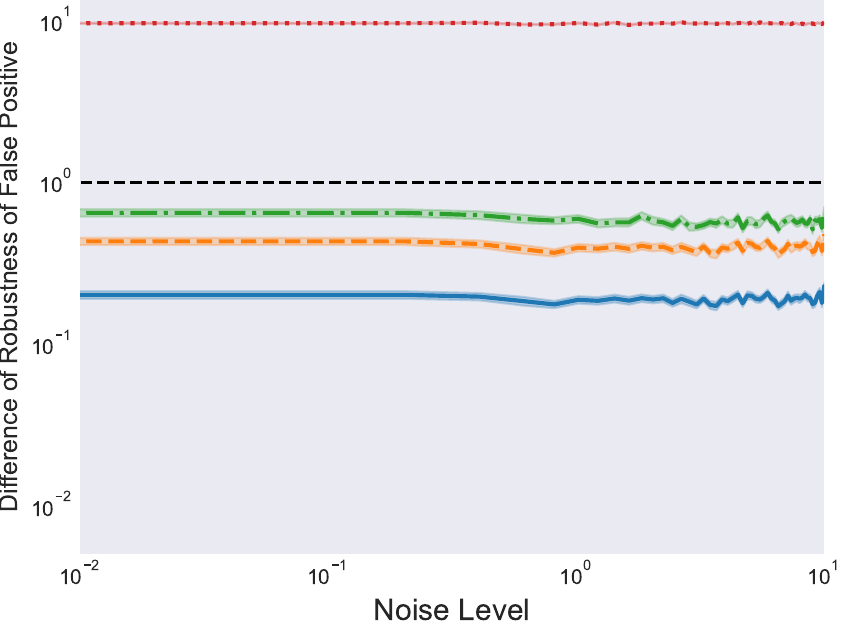}
         \caption{False Positive robustness.}
         \label{bank_rob_fp}
     \end{subfigure}
     \begin{subfigure}[b]{0.475\textwidth}
         \centering
         \includegraphics[width=.875\textwidth]{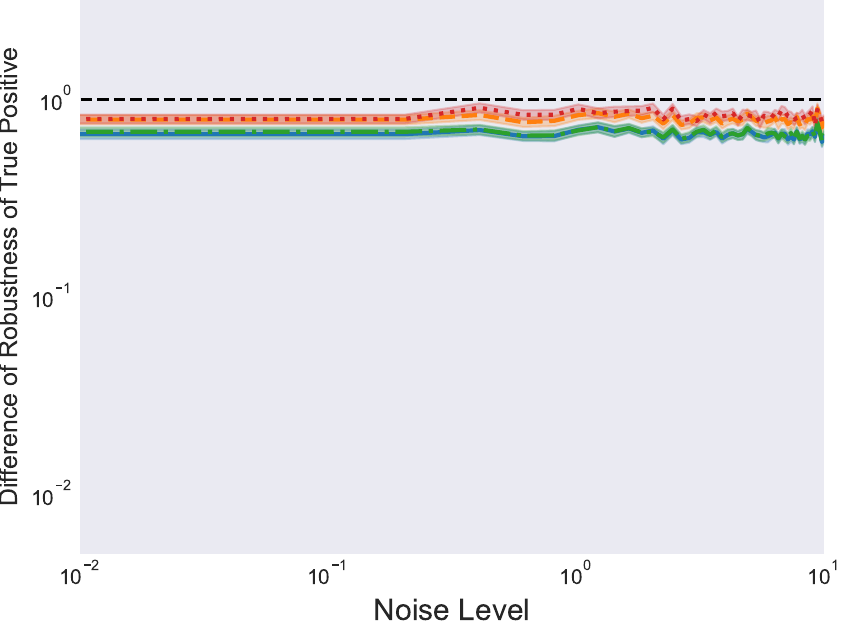}
         \caption{True Positive robustness.}
         \label{bank_rob_tp}
     \end{subfigure}
         \end{multicols}
          \caption{Fairness and robustness of fairness for various strategies using the  Bank Marketing data set with na\"ive Bayes.}
     \label{bank}
\end{figure}

\paragraph{Fairness -- Bank Marketing}

The evaluation for the fairness of the Bank Marketing data set is found in Figure~\ref{bank}(\subref{bank_fair_dp}, \subref{bank_fair_eo}, \subref{bank_fair_fp} \& \subref{bank_fair_tp}). Again, we observe the same behaviour for demographic parity, equalised odds, and false positive rate as previous data sets -- increasing fairness for all methods except post-processing. Performance for the true positive rate is almost indistinguishable between methods.

\paragraph{Robustness -- Bank Marketing}

The robustness behaviour for the Bank Marketing data set can be found in Figure~\ref{bank}(\subref{bank_rob_dp}, \subref{bank_rob_eo}, \subref{bank_rob_fp} \& \subref{bank_rob_tp}). Despite matching the general behaviour found in previous data sets, we can see that the instability seems to occur unusually early for the post-processing method. In fact, the result shows that
    \begin{equation*}
        \frac{\partial R_k(f_{post}, M, \mathbf{X})}{\partial k}\Bigg\vert_{k\in[2,10]}\in\Bigg[-\frac{3}{500},\frac{1}{1000}\Bigg]
    \end{equation*} 
for all metrics except true positive. Having a small average rate of change in the robustness for low noise means that the model finds stability quite quickly. However, this stability is still unsatisfactory for the post-processing method as $R_k > 1$ for $k>0$, so the fairness is worse for noisy data.

\subsubsection{German Credit}

The results for the German Credit data set are restricted to SVM, with K=100. We choose this method for the German Credit data set because it contains the smallest number of samples, and SVM is the most computationally intensive method to use.

\begin{figure}
         \begin{multicols}{2}
     \centering
     \begin{subfigure}[b]{0.475\textwidth}
         \centering
         \includegraphics[width=.875\textwidth]{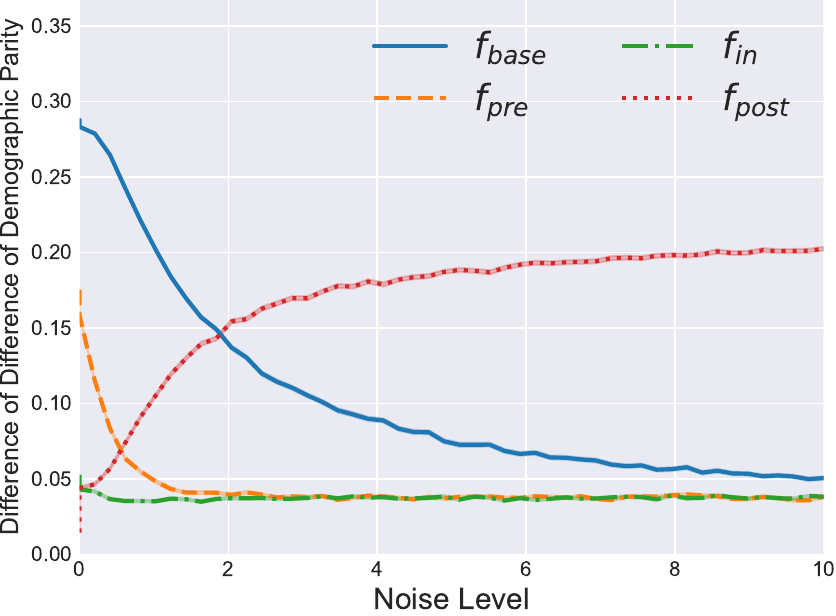}
         \caption{Demographic Parity fairness.}
         \label{german_fair_dp}
     \end{subfigure}
     \begin{subfigure}[b]{0.475\textwidth}
         \centering
         \includegraphics[width=.875\textwidth]{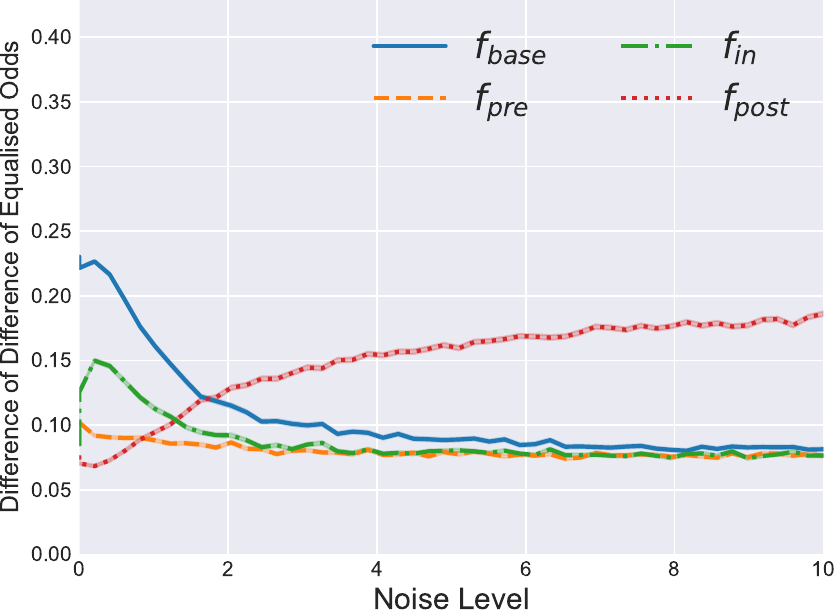}
         \caption{Equalised Odds fairness.}
         \label{german_fair_eo}
     \end{subfigure}
     \begin{subfigure}[b]{0.475\textwidth}
         \centering
         \includegraphics[width=.875\textwidth]{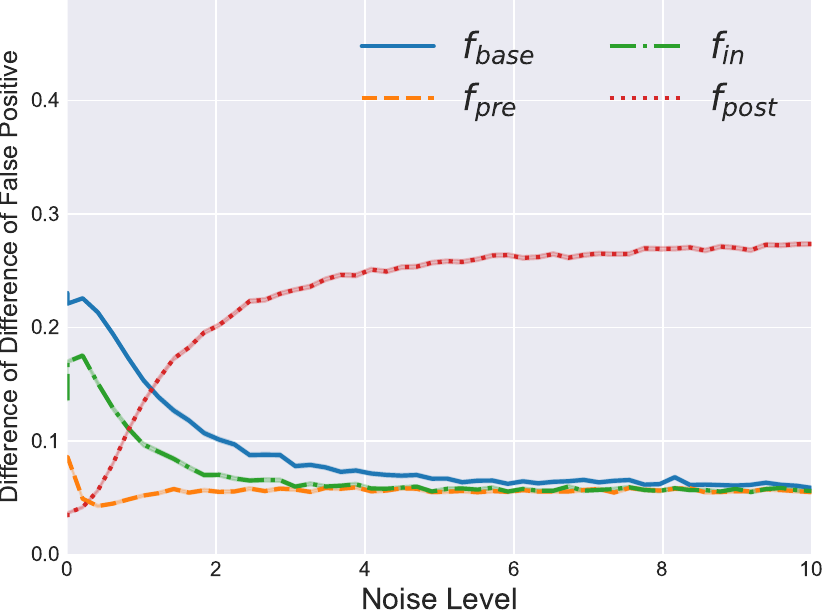}
         \caption{False Positive fairness.}
         \label{german_fair_fp}
     \end{subfigure}
     \begin{subfigure}[b]{0.475\textwidth}
         \centering
         \includegraphics[width=.875\textwidth]{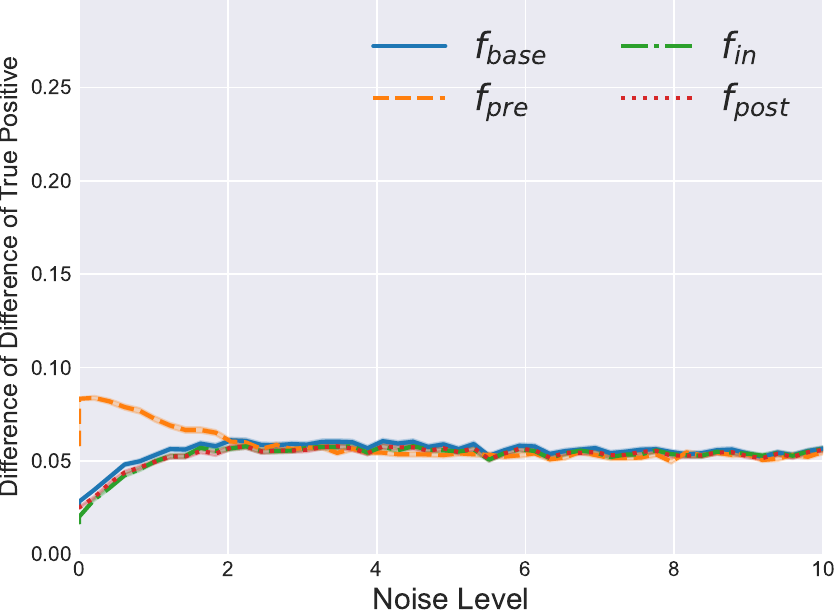}
         \caption{True Positive fairness.}
         \label{german_fair_tp}
     \end{subfigure}
     \begin{subfigure}[b]{0.475\textwidth}
         \centering
         \includegraphics[width=.875\textwidth]{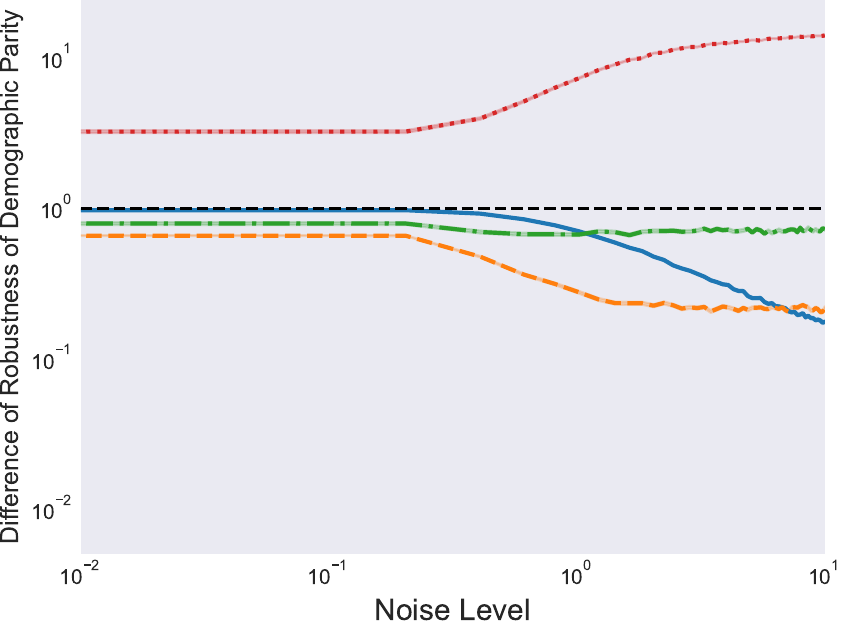}
         \caption{Demographic Parity robustness.}
         \label{german_rob_dp}
     \end{subfigure}
     \begin{subfigure}[b]{0.475\textwidth}
         \centering
         \includegraphics[width=.875\textwidth]{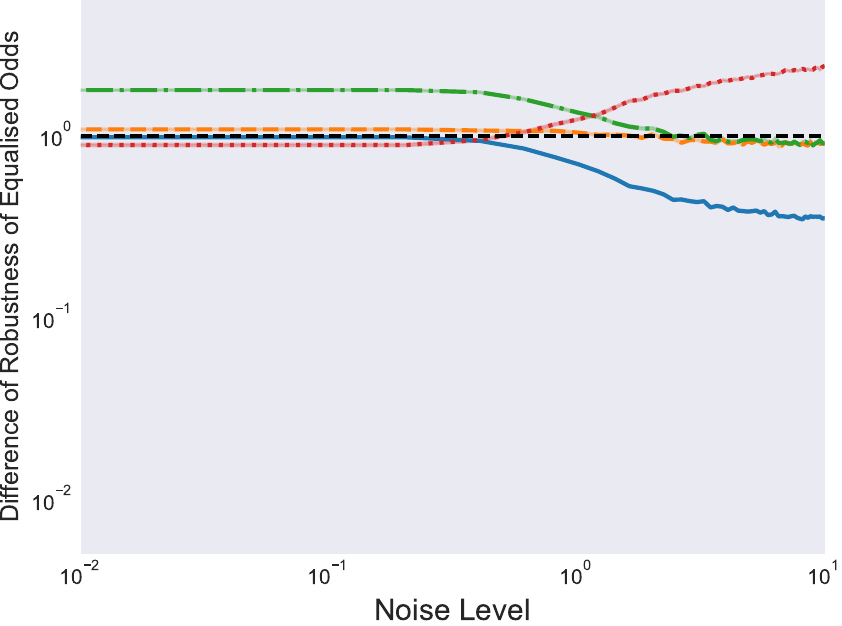}
         \caption{Equalised Odds robustness.}
         \label{german_rob_eo}
     \end{subfigure}
     \begin{subfigure}[b]{0.475\textwidth}
         \centering
         \includegraphics[width=.875\textwidth]{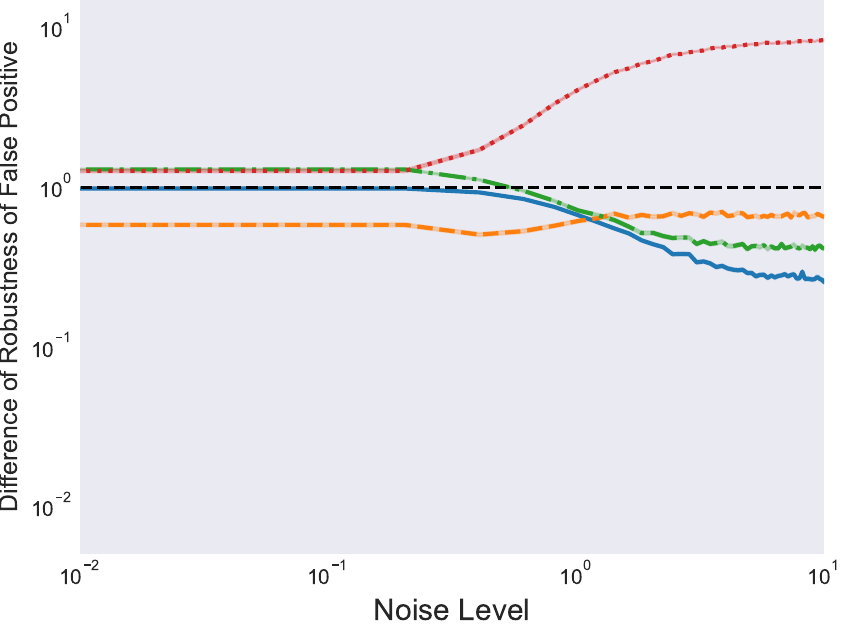}
         \caption{False Positive robustness.}
         \label{german_rob_fp}
     \end{subfigure}
     \begin{subfigure}[b]{0.475\textwidth}
         \centering
         \includegraphics[width=.875\textwidth]{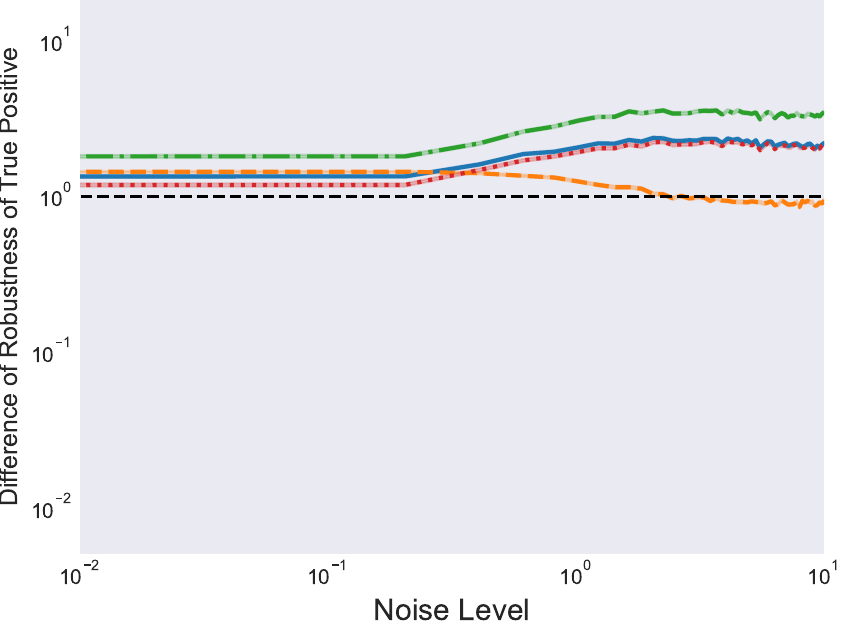}
         \caption{True Positive robustness.}
         \label{german_rob_tp}
     \end{subfigure}
         \end{multicols}
          \caption{Fairness and robustness of fairness for various strategies using the German Credit data set with support vector machines.}
     \label{german}
\end{figure}

\paragraph{Fairness -- German Credit}

The fairness measures for noisy data with the German Credit data set are displayed in Figure~\ref{german}(\subref{german_fair_dp}, \subref{german_fair_eo}, \subref{german_fair_fp} \& \subref{german_fair_tp}). Notably, the true positive rate for $k=0$ is very poor for $f_{pre}$ when compared to other methods -- it even performs worse than the baseline $f_{base}$. This could be due to the class imbalance between the protected classes -- the linear transformation removes the dependency the data has on $\mathbf{A}=1$ but not $\mathbf{A}=0$.

\paragraph{Robustness -- German Credit}

The robustness metric for noisy German Credit data is shown in Figure~\ref{german}(\subref{german_rob_dp}, \subref{german_rob_eo}, \subref{german_rob_fp} \& \subref{german_rob_tp}). While we do see the usual trend of the post-processing method wandering away from its initial solution  as $k$ increases, suprisingly we also see poor robustness across the board for the true positive rate. More specifically:
\begin{itemize}
    \item $R_k(M_{tp}, f_{pre}, \mathbf{X})>1$ for small values of $k$; and
    \item $R_k(M_{tp}, f_{\{base,in,post\}},\mathbf{X}) > 1$ for all $k$.
\end{itemize}
Again, this is likely due to two things. Firstly, the class ratio for the target data is poorly balanced -- people are almost three times as likely to have a bad credit rating than a good one. Adding to this, the ratio in the protected class is also poor. As such, optimising for the true positive rate is more challenging, as a smaller portion of the data is classed as positive. Only 10.9\% of the data are classed as $\{\mathbf{A}=0, \mathbf{Y}=1\}$, and so the linear transformation does not have much information. This could also be the cause of the instability in $f_{in}$ in both equalised odds and false positive for small values of $k$.

\subsubsection{Law School}

The fairness behaviour for noisy Law School data is restricted to LR, with $K=100$. We choose to use logistic regression because the feature size is small, with no non-meaningful data.

\begin{figure}
         \begin{multicols}{2}
     \centering
     \begin{subfigure}[b]{0.475\textwidth}
         \centering
         \includegraphics[width=.875\textwidth]{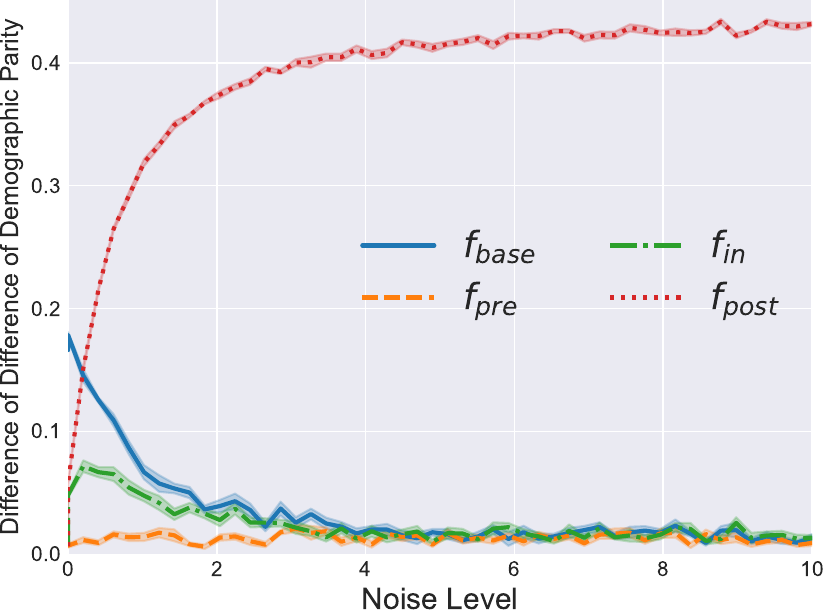}
         \caption{Demographic Parity fairness.}
         \label{law_fair_dp}
     \end{subfigure}
     \begin{subfigure}[b]{0.475\textwidth}
         \centering
         \includegraphics[width=.875\textwidth]{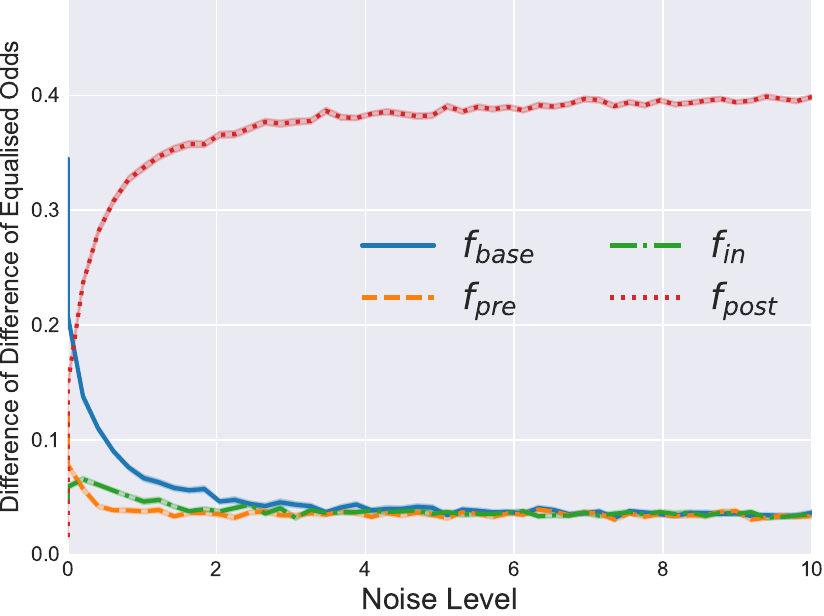}
         \caption{Equalised Odds fairness.}
         \label{law_fair_eo}
     \end{subfigure}
     \begin{subfigure}[b]{0.475\textwidth}
         \centering
         \includegraphics[width=.875\textwidth]{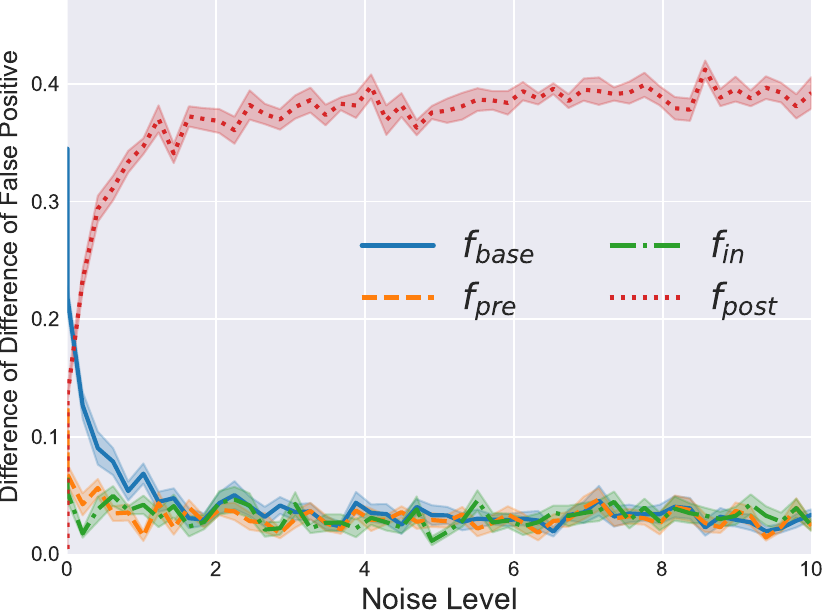}
         \caption{False Positive fairness.}
         \label{law_fair_fp}
     \end{subfigure}
     \begin{subfigure}[b]{0.475\textwidth}
         \centering
         \includegraphics[width=.875\textwidth]{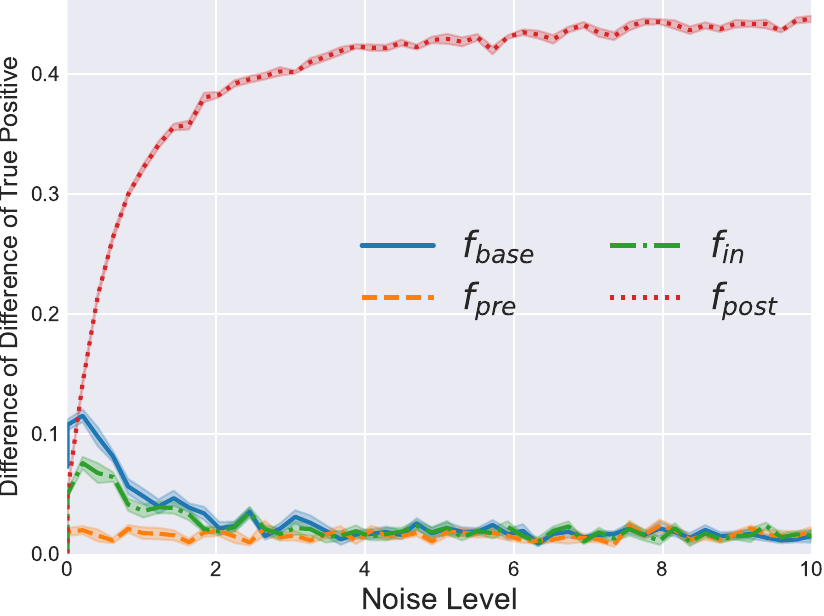}
         \caption{True Positive fairness.}
         \label{law_fair_tp}
     \end{subfigure}
     \begin{subfigure}[b]{0.475\textwidth}
         \centering
         \includegraphics[width=.875\textwidth]{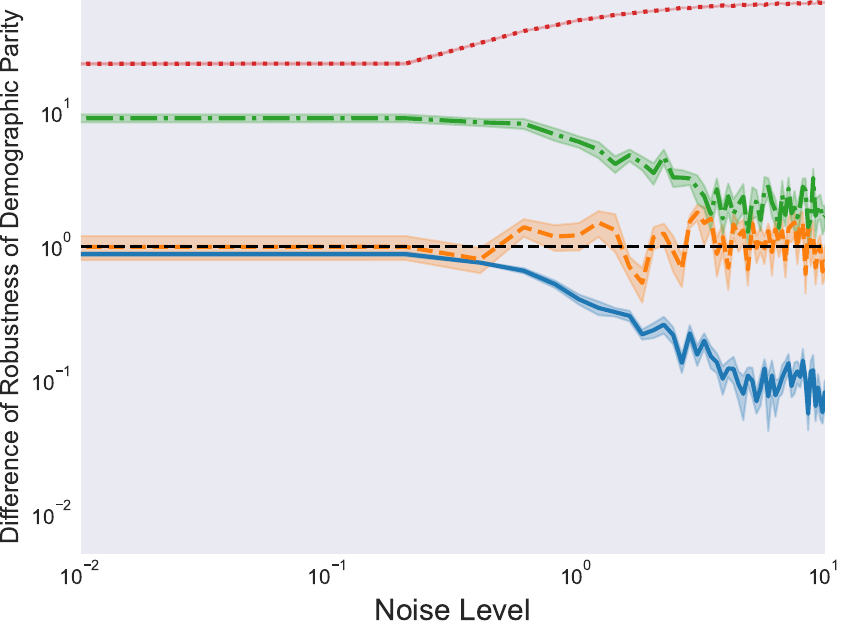}
         \caption{Demographic Parity robustness.}
         \label{law_rob_dp}
     \end{subfigure}
     \begin{subfigure}[b]{0.475\textwidth}
         \centering
         \includegraphics[width=.875\textwidth]{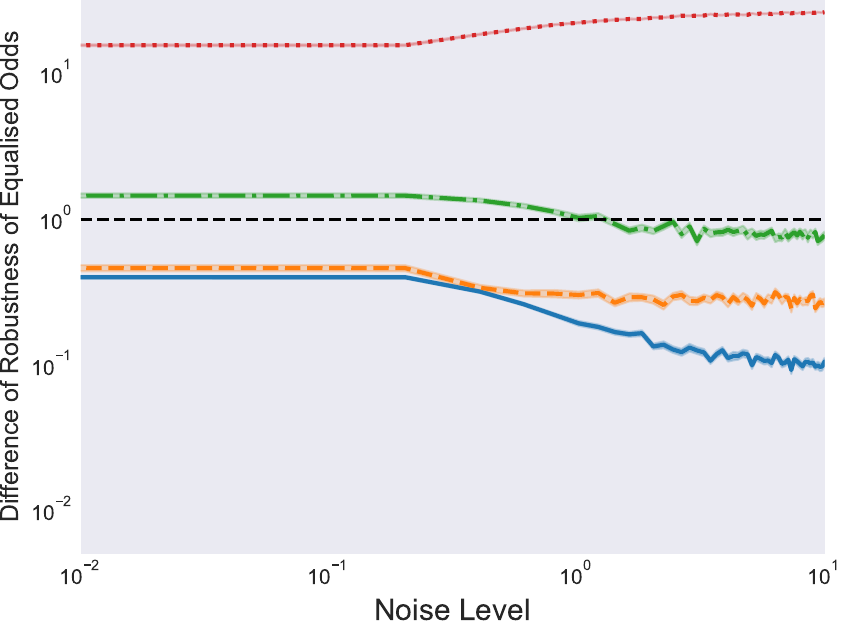}
         \caption{Equalised Odds robustness.}
         \label{law_rob_eo}
     \end{subfigure}
     \begin{subfigure}[b]{0.475\textwidth}
         \centering
         \includegraphics[width=.875\textwidth]{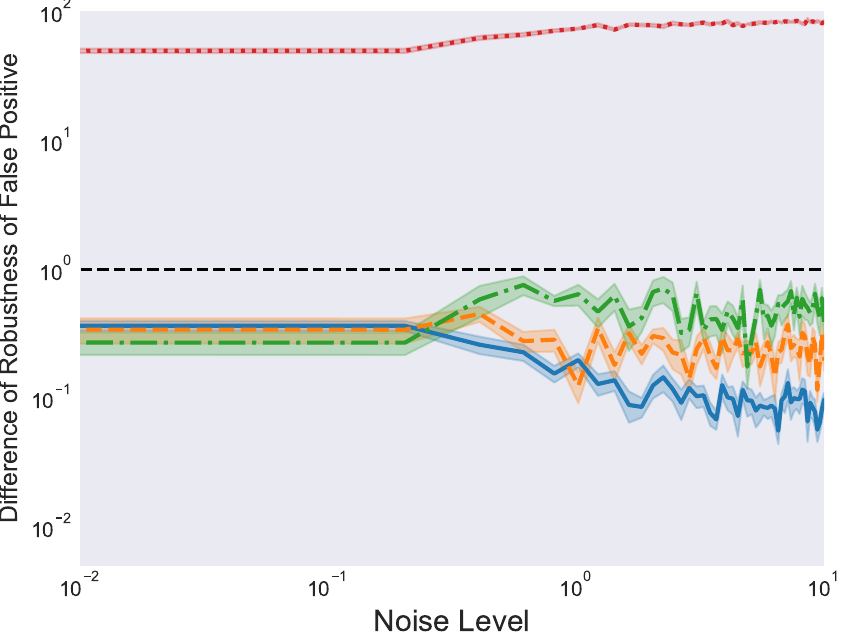}
         \caption{False Positive robustness.}
         \label{law_rob_fp}
     \end{subfigure}
     \begin{subfigure}[b]{0.475\textwidth}
         \centering
         \includegraphics[width=.875\textwidth]{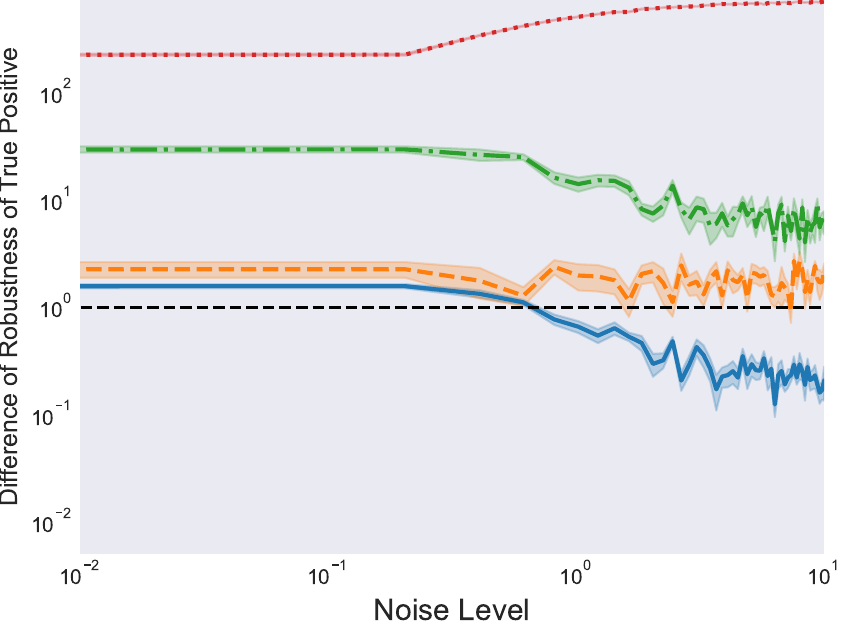}
         \caption{True Positive robustness.}
         \label{law_rob_tp}
     \end{subfigure}
         \end{multicols}
          \caption{Fairness and robustness of fairness for various strategies using the Law School data set with logistic regression.}
     \label{law}
\end{figure}

\paragraph{Fairness -- Law School}

The results in Figure~\ref{law}(\subref{law_fair_dp}, \subref{law_fair_eo}, \subref{law_fair_fp} \& \subref{law_fair_tp}) follow the findings discovered for the other data sets. However, we do see a particularly poor performance in equalised odds and false positive for $f_{base}$. This is likely caused by a strong bias inherent in the data.
Since equalised odds is combination of the false and true positive constraints, it logically follows that a poor performance for $f_{base}$ in false positive would cause a poor performance in equalised odds.

\paragraph{Robustness -- Law School}

The robustness for the LSAC data is show in Figure~\ref{law}(\subref{law_rob_dp}, \subref{law_rob_eo}, \subref{law_rob_fp} \& \subref{law_rob_tp}). Both the baseline and pre-processing method offer good robustness for this data set, but the in-processing method $f_{in}$ misses the mark in true positive rate and demographic parity. Again, we see that $R_k(M, f_{in}, \mathbf{X})>1$ for small values of $k$ for some fairness metrics. This behaviour is particularly interesting, as we would expect some amount of instability for large values of $k$. However, small fluctuations to the data set should not cause such large relative shifts in fairness.
Notably, we see that  $R_k(M, f_{post}, \mathbf{X}) > 1$ for all metrics. This severe instability of $f_{post}$ has been a consistent theme throughout the experimentation. Because of this, we must assume that the mechanisms directly behind threshold optimisation are the underlying cause of the instability. We investigate this phenomenon in more detail in Section~\ref{sec:disc}.

\subsection{Summary}

As we add stronger noise to the data sets, the model's ability to make accurate predictions weakens. However, our experiments show that errors are not proportioned fairly between protected subgroups. On the contrary, in some scenarios, one group is significantly advantaged over the other, such as in the Law School example.
%
We can see that the ability of a model to allocate errors in a manner perceived as \emph{fair} -- according to the chosen notion of fairness -- depends on the fairness definition, the learning model, and the data set. While more experiments are warranted, we clearly see two emerging patterns:
\begin{enumerate}
    \item equalised odds and false positive rate can have instabilities for small values of $k$; and
    \item the threshold optimisation method  shows a particular inability to share errors fairly among protected groups.
\end{enumerate}

\section{Analysis of Threshold Optimisation\label{sec:disc}}

The experimental results show that the post-processing method \textit{threshold optimisation} demonstrates a different behaviour to the other three methods. More specifically, it seems to get less fair as the noise level increases, whereas other methods often show improvement. 
As such, it is important to analyse why this is the case. To this end, we explore the post-processing method under the equalised odds criterion for noisy protected data, as shown by \citet{eotheory}, and find the maximum expected unfairness for demographic parity. Here, we refer to the pre-fitted model as $f(\mathbf{X})$ with the data set $\mathbf{X}$, and the threshold optimisation model on $f$ being $\mathbf{F}(f(\mathbf{X}))$.
As mentioned before, we add noise to the protected class $\mathbf{A}$ to create $\mathbf{A}_c$. As such, $\mathbf{A}=a$ represents the true class, and $\mathbf{A}_c=a_c$ is a potential incorrect entry of $\mathbf{A}$.

\begin{definition}
The bias for class $\mathbf{Y}=y$ of the predictor $\mathbf{F}$ is defined in the same way as Definition~\ref{EO}, but we now specify the value of $y$ to differentiate between false positives and true positives:
\begin{equation*}
\begin{aligned}
    &M_{eo}(\mathbf{F},\mathbf{X})\big\vert_{\mathbf{Y}=y}\\
    &=\left \vert\mathbb{P}\{\mathbf{F}(\boldsymbol{X})=1\vert\mathbf{Y}=\boldsymbol{y},\mathbf{A}=0\} 
    - \mathbb{P}\{\mathbf{F}(\boldsymbol{X})=1\vert\mathbf{Y}=\boldsymbol{y},\mathbf{A}=1\}\right\vert
    \text{~.}
    \end{aligned}
\end{equation*}
\end{definition}

\subsection{Bounds on Fairness}

\begin{theorem}[Bound of Bias under Equalised Odds]
    Let $\mathbf{Y}$ be a sample's ground truth and $\mathbf{A}$ be its true protected attribute. To distinguish between functions, let $\mathbf{F}:f(\mathcal{X})\mapsto[0,1]$ be the post-processing equalised odds predictor from the pre-fitted classifier $f$. If we inject noise of strength $k\in[0,1]$ (which we refer to as the Bernoulli noise), where $k=1$ is the maximum noise, then the bias is precisely bounded by
    \begin{equation}
    \begin{aligned}
        M_{eo}(\mathbf{F},\mathbf{X})\big\vert_{\mathbf{Y}=1}&\leq \alpha M_{eo}(f,\mathbf{X})\big\vert_{\mathbf{Y}=1} \\
         M_{eo}(\mathbf{F},\mathbf{X})\big\vert_{\mathbf{Y}=1}&\geq M_{eo}(f,\mathbf{X})\big\vert_{\mathbf{Y}=1}(1-k)(p_{1,0}-p_{0,0}+p_{1,1}-p_{0,1})
         \text{~,}
         \label{bounds}
    \end{aligned}
    \end{equation}
    where $p_{y, a}$ denotes the probability that $\mathbf{F}(f(\mathbf{X}))=1$ given $f(\mathbf{X})=y$ and $\mathbf{A}=a$, and
    \begin{equation*}
        \alpha = \left\vert\frac{p_{1,0}-p_{0,0}}{\mathbb{P}\{\mathbf{Y}=1,\mathbf{A}_c=0\}}\right\vert
        +\left\vert\frac{p_{1,1}-p_{0,1}}{\mathbb{P}\{\mathbf{Y}=1,\mathbf{A}_c=1\}}\right\vert
        \text{~.}
    \end{equation*}
As such, the difference for equalised odds is dependent on both the strength of the noise injection and the difference between the probabilities used in the model. (Proof in Appendix~\ref{apx:bias_bound}.)
\label{th:bound}
\end{theorem}

\begin{theorem}[Maximum Demographic Parity due to Noise]
    Given a model $f:\mathcal{X}\mapsto\mathcal{Y}$ and a threshold optimiser $\mathbf{F}:f(\mathcal{X})\mapsto[0,1]$, the unfairness limit due to noise is given by 
    \begin{equation}
    \begin{aligned}
        M_{dp}(F,\mathbf{\tilde{X}}_\infty) = \big\vert&(p_{1,0}\mathbb{P}\{f(\mathbf{\tilde{X}}_\infty)\geq T_{1,0}\} + p_{0,0}\mathbb{P}\{f(\mathbf{\tilde{X}}_\infty)\geq T_{0,0}\}) \\- &(p_{0,1}\mathbb{P}\{f(\mathbf{\tilde{X}}_\infty)\geq T_{0,1}\} + p_{1,1}\mathbb{P}\{f(\mathbf{\tilde{X}}_\infty)\geq T_{1,1}\})\big\vert
        \text{~,}
        \end{aligned}
        \label{maxrand}
    \end{equation}
where $\tilde{X}_\infty$ is very noisy data. This result also holds for a single threshold, where we simply consider $T_{1,a} = T_{0,a}$.
(Proof in Appendix~\ref{apx:max}.)
\label{th:max}
\end{theorem}

Again, the maximum demographic parity due to noise is a logical conclusion. As the data become noisier, the signals within the data are destroyed. Therefore, the binary classifier $F$ becomes increasingly trivial (i.e., it must guess for more and more inputs). At maximum noise, or completely random data, the output distribution becomes uniform. Thus, the difference in output is then purely decided by the difference between the thresholds and probabilities for each protected group as shown in Equation~\ref{maxrand}.

\section{Conclusion and Future Research}
In this paper, we investigated the robustness and stability of fair optimisation strategies. We proposed a new way to measure the robustness, named \textit{relative robustness}, and explored how the different strategies offer different levels of robustness. In our experimentation, we demonstrated that large perturbations to the data for the \textit{exponentiated gradient} in-learning method show very little impact on the fairness of the model, and that the post-processing method, though bounded by Theorem~\ref{th:max}, is particularly sensitive to noise. We also showed that using a linear pre-processing method to remove bias is not always an appropriate method to rid a data set of bias.
We have also proposed a novel framework to explore the robustness of fair models, which allows us to compare the stability of different solutions with respect to fairness. This work makes a significant contribution towards ensuring that our data-driven models continue to make fair decisions beyond deployment, and can assist in predicting how and when a model's outputs may become unfair.

While this work has offered us some answers, it has also opened new avenues of research. Specific lines of inquiry include:
\begin{enumerate}
    \item Do other pre/in/post-processing methods offer a similar kind of robustness?
    \item Is the robustness linked more heavily to the learning model or the fairness constraint location? 
    \item Can Theorem~\ref{th:bound} be extended to find a bound on fairness where all data have injected noise?
    \item Can we further generalise the result from Theorem~\ref{th:max}, expanding it beyond equalised odds to other fairness constraints?
    \item How robust are the methods under more specific changes to distribution, not just random noise?
\end{enumerate}

In future work, we plan to conduct more thorough and in-depth experiments into each of the fairness strategies (pre-processing, in-processing, and post-processing). Furthermore, we wish to engage with more learning models, as well as introduce other notions of fairness and explore problems beyond binary classification. Some methods also demonstrate severe instability for small noise injections -- see the false positive rate for the Adult Income data set in Figure~\ref{adult}(\subref{adult_fair_fp} \& \subref{adult_rob_fp}), true positive rate for the Law School data set in Figure~\ref{law}(\subref{law_fair_tp} \& \subref{law_rob_tp}), or the true positive rate for the German Credit data set in Figure~\ref{german}(\subref{german_fair_tp} \& \subref{german_rob_tp}). This is particularly concerning, as small variations happening in the input space are easy to miss, and can even occur between training and validation data, as shown by \citet{https://doi.org/10.48550/arxiv.2007.06029}.

We also limited our exploration to binary classification. This was purely for practical purposes, as it allowed us to explore other aspects of fairness strategies in greater detail for simple problems. However, most machine learning models are now deployed in more complex areas -- input spaces are of higher dimensionality, and classification is significantly more diverse and complicated. An exploration into the robustness of these areas would extend the impact of our findings, especially as we can refer back to the results reported here.

\section*{Declarations}


\paragraph{Funding}


This research was supported by the ARC Centre of Excellence for Automated Decision-Making and Society, funded by the Australian Government through the Australian Research Council (project number CE200100005). 


\paragraph{Conflicts of Interest}
The authors declare no competing interests.%

\paragraph{Data Transparency}
The following data sets were used in our research:
\begin{description}
    \item 
    [Adult Income]
\url{https://github.com/zykls/folktables} %
    \item 
    [Bank Marketing]
\url{https://archive.ics.uci.edu/ml/datasets/bank+marketing} %
    \item 
    [COMPAS]
\url{https://github.com/propublica/compas-analysis} %
    \item 
    [German Credit]
\url{https://archive.ics.uci.edu/ml/datasets/statlog+(german+credit+data)} %
    \item 
    [Law School]
\url{https://eric.ed.gov/?id=ED469370} %
\end{description}
They are benchmark data sets that are freely available online.

\paragraph{Code Availability}
\url{https://github.com/TeddyZander/FairR}

\paragraph{Authors' Contributions}
\begin{description}
    \item [Conceptualization:] Edward A.\ Small, Wei Shao, Flora D.\ Salim.%
    \item [Methodology:] Edward A.\ Small, Kacper Sokol.%
    \item [Formal analysis and investigation:] Edward A.\ Small.%
    \item [Contributions to formal analysis and investigation:] Zeliang Zhang, Peihan Liu.%
    \item [Writing -- original draft preparation:] Edward A.\ Small. %
    \item [Writing -- review and editing:] Edward A.\ Small, Kacper Sokol, Wei Shao, Jeffrey Chan, Flora D.\ Salim.%
    \item [Funding acquisition:] Jeffrey Chan, Flora D.\ Salim. %
    \item [Resources:] Edward A.\ Small. %
    \item [Supervision:] Kacper Sokol, Jeffrey Chan, Flora D.\ Salim.%
\end{description}

\bibliography{bib}%

\begin{thebibliography}{56}
\providecommand{\natexlab}[1]{#1}
\providecommand{\url}[1]{{#1}}
\providecommand{\urlprefix}{URL }
\expandafter\ifx\csname urlstyle\endcsname\relax
  \providecommand{\doi}[1]{DOI~\discretionary{}{}{}#1}\else
  \providecommand{\doi}{DOI~\discretionary{}{}{}\begingroup
  \urlstyle{rm}\Url}\fi
\providecommand{\eprint}[2][]{\url{#2}}

\bibitem[{Agarwal et~al.(2018)Agarwal, Beygelzimer, Dud{\'\i}k, Langford, and
  Wallach}]{https://doi.org/10.48550/arxiv.1803.02453}
Agarwal A, Beygelzimer A, Dud{\'\i}k M, Langford J, Wallach H (2018) A
  reductions approach to fair classification. In: International conference on
  machine learning, PMLR, pp 60--69

\bibitem[{Ahn and Lin(2019)}]{ahn2019fairsight}
Ahn Y, Lin YR (2019) Fairsight: {Visual} analytics for fairness in decision
  making. IEEE Transactions on Visualization and Computer Graphics
  26(1):1086--1095

\bibitem[{Anderson and Sojoudi(2020)}]{anderson2020certifying}
Anderson BG, Sojoudi S (2020) Certifying neural network robustness to random
  input noise from samples. arXiv preprint arXiv:201007532

\bibitem[{Awasthi et~al.(2020)Awasthi, Kleindessner, and
  Morgenstern}]{eotheory}
Awasthi P, Kleindessner M, Morgenstern J (2020) Equalized odds postprocessing
  under imperfect group information. In: International conference on artificial
  intelligence and statistics, PMLR, pp 1770--1780

\bibitem[{Barenstein(2019)}]{compas}
Barenstein M (2019) {ProPublica}'s {COMPAS} data revisited. arXiv preprint
  arXiv:190604711

\bibitem[{Bellamy et~al.(2018)Bellamy, Dey, Hind, Hoffman, Houde, Kannan,
  Lohia, Martino, Mehta, Mojsilovic, Nagar, Ramamurthy, Richards, Saha,
  Sattigeri, Singh, Varshney, and Zhang}]{aif360-oct-2018}
Bellamy RKE, Dey K, Hind M, Hoffman SC, Houde S, Kannan K, Lohia P, Martino J,
  Mehta S, Mojsilovic A, Nagar S, Ramamurthy KN, Richards J, Saha D, Sattigeri
  P, Singh M, Varshney KR, Zhang Y (2018) {AI} {Fairness} 360: {An} extensible
  toolkit for detecting, understanding, and mitigating unwanted algorithmic
  bias. arXiv preprint arXiv:181001943

\bibitem[{Bird et~al.(2020)Bird, Dud{\'\i}k, Edgar, Horn, Lutz, Milan, Sameki,
  Wallach, and Walker}]{bird2020fairlearn}
Bird S, Dud{\'\i}k M, Edgar R, Horn B, Lutz R, Milan V, Sameki M, Wallach H,
  Walker K (2020) Fairlearn: {A} toolkit for assessing and improving fairness
  in {AI}. Microsoft, Tech Rep MSR-TR-2020-32

\bibitem[{Biswas and Rajan(2021)}]{biswas2021fair}
Biswas S, Rajan H (2021) Fair preprocessing: {Towards} understanding
  compositional fairness of data transformers in machine learning pipeline. In:
  Proceedings of the 29\textsuperscript{th} ACM joint meeting on European
  software engineering conference and symposium on the foundations of software
  engineering, pp 981--993

\bibitem[{Calmon et~al.(2017)Calmon, Wei, Vinzamuri, Natesan~Ramamurthy, and
  Varshney}]{calmon2017optimized}
Calmon F, Wei D, Vinzamuri B, Natesan~Ramamurthy K, Varshney KR (2017)
  Optimized pre-processing for discrimination prevention. Advances in Neural
  Information Processing Systems 30:3995--4004

\bibitem[{Carlini and Wagner(2017)}]{carlini2017towards}
Carlini N, Wagner D (2017) Towards evaluating the robustness of neural
  networks. In: 2017 IEEE Symposium on Security and Privacy (SP), IEEE, pp
  39--57

\bibitem[{Carrara et~al.(2017)Carrara, Falchi, Caldelli, Amato, Fumarola, and
  Becarelli}]{carrara2017detecting}
Carrara F, Falchi F, Caldelli R, Amato G, Fumarola R, Becarelli R (2017)
  Detecting adversarial example attacks to deep neural networks. In:
  Proceedings of the 15\textsuperscript{th} International Workshop on
  Content-Based Multimedia Indexing, pp 1--7

\bibitem[{Castelnovo et~al.(2021)Castelnovo, Malandri, Mercorio, Mezzanzanica,
  and Cosentini}]{10.1007/978-3-030-93736-2_46}
Castelnovo A, Malandri L, Mercorio F, Mezzanzanica M, Cosentini A (2021)
  Towards fairness through time. In: Joint European Conference on Machine
  Learning and Knowledge Discovery in Databases, Springer, pp 647--663

\bibitem[{Castelnovo et~al.(2022)Castelnovo, Crupi, Greco, Regoli, Penco, and
  Cosentini}]{group}
Castelnovo A, Crupi R, Greco G, Regoli D, Penco IG, Cosentini AC (2022) A
  clarification of the nuances in the fairness metrics landscape. Scientific
  Reports 12(1):4209

\bibitem[{Cui et~al.(2021)Cui, Pan, Zhang, and Wang}]{cui2020xorder}
Cui S, Pan W, Zhang C, Wang F (2021) Towards model-agnostic post-hoc adjustment
  for balancing ranking fairness and algorithm utility. In: Proceedings of the
  27\textsuperscript{th} ACM SIGKDD Conference on Knowledge Discovery and Data
  Mining, pp 207--217

\bibitem[{Daxberger et~al.(2021)Daxberger, Kristiadi, Immer, Eschenhagen,
  Bauer, and Hennig}]{daxberger2021laplace}
Daxberger E, Kristiadi A, Immer A, Eschenhagen R, Bauer M, Hennig P (2021)
  Laplace redux-effortless bayesian deep learning. Advances in Neural
  Information Processing Systems 34

\bibitem[{Deng et~al.(2012)Deng, Tian, and Zhang}]{SVM}
Deng N, Tian Y, Zhang C (2012) Support vector machines: {Optimization} based
  theory, algorithms, and extensions. CRC Press, New York

\bibitem[{Ding et~al.(2021)Ding, Hardt, Miller, and
  Schmidt}]{https://doi.org/10.48550/arxiv.2108.04884}
Ding F, Hardt M, Miller J, Schmidt L (2021) Retiring adult: {New} datasets for
  fair machine learning. Advances in Neural Information Processing Systems
  34:6478--6490

\bibitem[{Du et~al.(2020)Du, Yang, Zou, and Hu}]{du2020fairness}
Du M, Yang F, Zou N, Hu X (2020) Fairness in deep learning: A computational
  perspective. IEEE Intelligent Systems 36(4):25--34

\bibitem[{Farokhi(2021)}]{farokhi2021optimal}
Farokhi F (2021) Optimal pre-processing to achieve fairness and its
  relationship with total variation barycenter. arXiv preprint arXiv:210106811

\bibitem[{Fawzi et~al.(2016)Fawzi, Moosavi-Dezfooli, and
  Frossard}]{fawzi2016robustness}
Fawzi A, Moosavi-Dezfooli SM, Frossard P (2016) Robustness of classifiers:
  {From} adversarial to random noise. Advances in Neural Information Processing
  Systems 29

\bibitem[{Forsyth(2018)}]{SGD}
Forsyth D (2018) Learning to classify. Probability and Statistics for Computer
  Science pp 253--279

\bibitem[{Hardt et~al.(2016)Hardt, Price, and
  Srebro}]{https://doi.org/10.48550/arxiv.1610.02413}
Hardt M, Price E, Srebro N (2016) Equality of opportunity in supervised
  learning. Advances in Neural Information Processing Systems 29

\bibitem[{Hendrycks and Dietterich(2018)}]{hendrycks2019benchmarking}
Hendrycks D, Dietterich T (2018) Benchmarking neural network robustness to
  common corruptions and perturbations. In: International Conference on
  Learning Representations

\bibitem[{Hristea(2012)}]{NB}
Hristea FT (2012) The na{\"\i}ve {Bayes} model for unsupervised word sense
  disambiguation: {Aspects} concerning feature selection. Springer Science \&
  Business Media, Berlin, Heidelberg

\bibitem[{Kamishima et~al.(2011)Kamishima, Akaho, and
  Sakuma}]{kamishima2011fairness}
Kamishima T, Akaho S, Sakuma J (2011) Fairness-aware learning through
  regularization approach. In: 2011 IEEE 11\textsuperscript{th} International
  Conference on Data Mining Workshops, IEEE, pp 643--650

\bibitem[{Kim et~al.(2019)Kim, Ghorbani, and Zou}]{kim2019multiaccuracy}
Kim MP, Ghorbani A, Zou J (2019) Multiaccuracy: {Black-box} post-processing for
  fairness in classification. In: Proceedings of the 2019 AAAI/ACM Conference
  on AI, Ethics, and Society, pp 247--254

\bibitem[{Kivinen and Warmuth(1997)}]{KIVINEN19971}
Kivinen J, Warmuth MK (1997) Exponentiated gradient versus gradient descent for
  linear predictors. Information and Computation 132(1):1--63

\bibitem[{Kohavi(1996)}]{kohavi1996scaling}
Kohavi R (1996) Scaling up the accuracy of naive-{Bayes} classifiers: {A}
  decision-tree hybrid. In: Proceedings of the Second International Conference
  on Knowledge Discovery and Data Mining, pp 202--207

\bibitem[{Kowsari et~al.(2019)Kowsari, Jafari~Meimandi, Heidarysafa, Mendu,
  Barnes, and Brown}]{methodsurvey}
Kowsari K, Jafari~Meimandi K, Heidarysafa M, Mendu S, Barnes L, Brown D (2019)
  Text classification algorithms: {A} survey. Information 10(4):150

\bibitem[{Kuragano and Yamaguchi(2007)}]{kuragano2007curve}
Kuragano T, Yamaguchi A (2007) Curve shape modification and fairness
  evaluation. In: International Design Engineering Technical Conferences and
  Computers and Information in Engineering Conference, vol 48078, pp 459--468

\bibitem[{Kwon et~al.(2019)Kwon, Yoon, and Park}]{kwon2019poster}
Kwon H, Yoon H, Park KW (2019) {POSTER}: {Detecting} audio adversarial example
  through audio modification. In: Proceedings of the 2019 ACM SIGSAC Conference
  on Computer and Communications Security, pp 2521--2523

\bibitem[{Le~Quy et~al.(2022)Le~Quy, Roy, Iosifidis, Zhang, and
  Ntoutsi}]{QuyTaiLe2021Asod}
Le~Quy T, Roy A, Iosifidis V, Zhang W, Ntoutsi E (2022) A survey on datasets
  for fairness-aware machine learning. Wiley Interdisciplinary Reviews: Data
  Mining and Knowledge Discovery 12(3):e1452

\bibitem[{Lohia et~al.(2019)Lohia, Ramamurthy, Bhide, Saha, Varshney, and
  Puri}]{lohia2019bias}
Lohia PK, Ramamurthy KN, Bhide M, Saha D, Varshney KR, Puri R (2019) Bias
  mitigation post-processing for individual and group fairness. In: 2019 IEEE
  international conference on acoustics, speech and signal processing (ICASSP),
  IEEE, pp 2847--2851

\bibitem[{Lopes et~al.(2019)Lopes, Yin, Poole, Gilmer, and
  Cubuk}]{lopes2019improving}
Lopes RG, Yin D, Poole B, Gilmer J, Cubuk ED (2019) Improving robustness
  without sacrificing accuracy with patch {Gaussian} augmentation. arXiv
  preprint arXiv:190602611

\bibitem[{Mandal et~al.(2020)Mandal, Deng, Jana, Wing, and
  Hsu}]{https://doi.org/10.48550/arxiv.2007.06029}
Mandal D, Deng S, Jana S, Wing J, Hsu DJ (2020) Ensuring fairness beyond the
  training data. Advances in Neural Information Processing Systems
  33:18445--18456

\bibitem[{Mehrabi et~al.(2021)Mehrabi, Morstatter, Saxena, Lerman, and
  Galstyan}]{bias}
Mehrabi N, Morstatter F, Saxena N, Lerman K, Galstyan A (2021) A survey on bias
  and fairness in machine learning. ACM Computing Surveys (CSUR) 54(6):1--35

\bibitem[{Minson et~al.(2018)Minson, VanEpps, Yip, and
  Schweitzer}]{MinsonJuliaA2018Ettt}
Minson JA, VanEpps EM, Yip JA, Schweitzer ME (2018) Eliciting the truth, the
  whole truth, and nothing but the truth: {The} effect of question phrasing on
  deception. Organizational Behavior and Human Decision Processes 147:76--93

\bibitem[{Nascimento et~al.(2022)Nascimento, Cavalcanti, and
  Da~Costa-Abreu}]{NASCIMENTO2022117032}
Nascimento FR, Cavalcanti GD, Da~Costa-Abreu M (2022) Unintended bias
  evaluation: {An} analysis of hate speech detection and gender bias mitigation
  on social media using ensemble learning. Expert Systems with Applications
  201:117032

\bibitem[{Omar et~al.(2022)Omar, Choi, Nyang, and Mohaisen}]{nlp}
Omar M, Choi S, Nyang D, Mohaisen D (2022) Robust natural language processing:
  {Recent} advances, challenges, and future directions. IEEE Access
  10:86038--86056

\bibitem[{Osborne(2014)}]{alma9921601010301341}
Osborne JW (2014) Best practices in logistic regression. Sage Publications,
  London

\bibitem[{Puga et~al.(2015)Puga, Krzywinski, and Altman}]{bayes}
Puga JL, Krzywinski M, Altman N (2015) Bayes' theorem: {Incorporate} new
  evidence to update prior information. Nature Methods 12(4):277--279

\bibitem[{Rajkomar et~al.(2018)Rajkomar, Hardt, Howell, Corrado, and
  Chin}]{rajkomar2018ensuring}
Rajkomar A, Hardt M, Howell MD, Corrado G, Chin MH (2018) Ensuring fairness in
  machine learning to advance health equity. Annals of Internal Medicine
  169(12):866--872

\bibitem[{Small et~al.(2024)Small, Sokol, Manning, Salim, and
  Chan}]{small2024equalised}
Small EA, Sokol K, Manning D, Salim FD, Chan J (2024) Equalised odds is not
  equal individual odds: {Post-processing} for group and individual fairness.
  In: Proceedings of the 2024 ACM Conference on Fairness, Accountability, and
  Transparency

\bibitem[{Sokol et~al.(2024)Sokol, Kull, Chan, and Salim}]{sokol2024ethical}
Sokol K, Kull M, Chan J, Salim F (2024) Cross-model fairness: {Empirical} study
  of fairness and ethics under model multiplicity. In: ACM Journal on
  Responsible Computing

\bibitem[{Song(2020)}]{SongM}
Song M (2020) Rethinking minority status and `visibility'. Comparative
  Migration Studies 8(1):5

\bibitem[{Tellinghuisen(1994)}]{TELLINGHUISEN1994255}
Tellinghuisen J (1994) On the least-squares fitting of correlated data:
  {Removing} the correlation. Journal of Molecular Spectroscopy 165(1):255--264

\bibitem[{Tomar et~al.(2021)Tomar, Shani, Efroni, and Ghavamzadeh}]{mirror}
Tomar M, Shani L, Efroni Y, Ghavamzadeh M (2021) Mirror descent policy
  optimization. In: International Conference on Learning Representations

\bibitem[{US(2011)}]{fairstandards}
US (2011) {The} {Equal} {Pay} {Act}. In: Code of Federal Regulations, Office of
  the Federal Register, USA, pp 56--57

\bibitem[{Wang et~al.(2020)Wang, Dong, Wang, Yan, and Wang}]{wang2020targeted}
Wang D, Dong L, Wang R, Yan D, Wang J (2020) Targeted speech adversarial
  example generation with generative adversarial network. IEEE Access
  8:124503--124513

\bibitem[{Wang and Zhai(2016)}]{DTC}
Wang X, Zhai J (2016) Learning with uncertainty. CRC Press, Boca Raton

\bibitem[{Wen et~al.(2021)Wen, Xu, Yang, He, and Huang}]{NEURIPS2021_28267ab8}
Wen P, Xu Q, Yang Z, He Y, Huang Q (2021) When false positive is intolerant:
  {End-to-end} optimization with low fpr for multipartite ranking. Advances in
  Neural Information Processing Systems 34:5025--5037

\bibitem[{Wojsznis et~al.(2007)Wojsznis, Mehta, and
  Thiele}]{DIRKTHIELE2007Antd}
Wojsznis WK, Mehta A, Thiele D (2007) Adding noise to data for model
  generation. Patent no.\ GB2437099A

\bibitem[{Xie et~al.(2012)Xie, Xu, and Chen}]{images}
Xie J, Xu L, Chen E (2012) Image denoising and inpainting with deep neural
  networks. Advances in Neural Information Processing Systems 25

\bibitem[{Yang et~al.(2017)Yang, Ren, Wang, and Dong}]{yang2017robust}
Yang L, Ren Z, Wang Y, Dong H (2017) A robust regression framework with
  {Laplace} kernel-induced loss. Neural Computation 29(11):3014--3039

\bibitem[{Yu et~al.(2019)Yu, Qin, Liu, Zhao, Wang, and
  Chen}]{yu2019interpreting}
Yu F, Qin Z, Liu C, Zhao L, Wang Y, Chen X (2019) Interpreting and evaluating
  neural network robustness. In: Proceedings of the 28\textsuperscript{th}
  International Joint Conference on Artificial Intelligence, pp 4199--4205

\bibitem[{Zafar et~al.(2017)Zafar, Valera, Rodriguez, Gummadi, and
  Weller}]{https://doi.org/10.48550/arxiv.1707.00010}
Zafar MB, Valera I, Rodriguez M, Gummadi K, Weller A (2017) From parity to
  preference-based notions of fairness in classification. Advances in Neural
  Information Processing Systems 30

\end{thebibliography}

\clearpage\newpage
\appendix

\section{Proofs}

\subsection{Fairness Convergence Under Noise\label{apx:convergence}}

    Take two distributions $p\sim N(\mu_p, \sigma^2_p)$ and $q\sim N(\mu_q, \sigma^2_q)$, which are parameterised by their mean $\mu$ and variance $\sigma$. Adding noise $\delta_k\sim N(0, k^2)$ to both distributions leads to
    \begin{equation*}
        \lim_{k\to\infty} D_B(p+\epsilon_k, q+\epsilon_k) = 0
        \text{~.}
    \end{equation*}

\begin{proof}
    From the definition of the normal distribution:
    \begin{equation*}
        \begin{aligned}
            p(x) = \frac{1}{\sigma_p\sqrt{2\pi}}\mathrm{e}^{-\frac{1}{2}\frac{x-\mu_p}{\sigma_p}} \\
            q(x) = \frac{1}{\sigma_q\sqrt{2\pi}}\mathrm{e}^{-\frac{1}{2}\frac{x-\mu_q}{\sigma_q}}
            \text{~.}
        \end{aligned}
    \end{equation*}
    As such:
    \begin{equation*}
        \begin{aligned}
            BC(p,q) &=\int_{-\infty}^\infty \sqrt{p(x)q(x)}dx \\
            &=\dfrac{\sqrt{2}\sqrt{{\sigma}_\text{p}}\sqrt{{\sigma}_\text{q}}}{\sqrt{{\sigma}_\text{q}^2+{\sigma}_\text{p}^2}}\mathrm{e}^{-\frac{{\mu}_\text{q}^2-2{\mu}_\text{p}{\mu}_\text{q}+{\mu}_\text{p}^2}{4{\sigma}_\text{q}^2+4{\sigma}_\text{p}^2}}
            \text{~.}
        \end{aligned}
    \end{equation*}
    Therefore:
    \begin{equation*}
    \begin{aligned}
        D_B(p,q) &= -\ln BC(p,q) \\
            &=-\ln\Bigg(\dfrac{\sqrt{2}\sqrt{{\sigma}_\text{p}}\sqrt{{\sigma}_\text{q}}}{\sqrt{{\sigma}_\text{q}^2+{\sigma}_\text{p}^2}}\mathrm{e}^{-\frac{{\mu}_\text{q}^2-2{\mu}_\text{p}{\mu}_\text{q}+{\mu}_\text{p}^2}{4{\sigma}_\text{q}^2+4{\sigma}_\text{p}^2}}\Bigg) \\
            & = \ln \Bigg( \mathrm{e}^{\frac{{\mu}_\text{q}^2-2{\mu}_\text{p}{\mu}_\text{q}+{\mu}_\text{p}^2}{4{\sigma}_\text{q}^2+4{\sigma}_\text{p}^2}}\Bigg)-\ln\Bigg( \dfrac{\sqrt{2}\sqrt{{\sigma}_\text{p}}\sqrt{{\sigma}_\text{q}}}{\sqrt{{\sigma}_\text{q}^2+{\sigma}_\text{p}^2}} \Bigg) \\
            &= \frac{1}{4}\frac{(\mu_p-\mu_q)^2}{\sigma_q^2+\sigma_p^2} + \ln\Bigg( \dfrac{\sqrt{{\sigma}_\text{q}^2+{\sigma}_\text{p}^2}}{\sqrt{2}\sqrt{{\sigma}_\text{p}}\sqrt{{\sigma}_\text{q}}} \Bigg) \\
            &= \frac{1}{4}\frac{(\mu_p-\mu_q)^2}{\sigma_q^2+\sigma_p^2} + \ln\Bigg(\sqrt{\frac{\sigma_q^2 + \sigma_p^2}{2\sigma_p\sigma_q}}\Bigg) \\
            &= \frac{1}{4}\frac{(\mu_p-\mu_q)^2}{\sigma_q^2+\sigma_p^2} + \frac{1}{2}\ln\Bigg(\frac{\sigma_q^2 + \sigma_p^2}{2\sqrt{\sigma_p^2}\sqrt{\sigma_q^2}}\Bigg)
            \text{~.}
    \end{aligned}
    \end{equation*}
    Adding noise $\delta_k\sim(0, k^2)$ to both distributions then gives:
    \begin{equation*}
        \begin{aligned}
            D_B(p+\delta_k,q+\delta_k) = \frac{1}{4}\frac{(\mu_p-\mu_q)^2}{\sigma_q^2+\sigma_p^2+2k^2} + \frac{1}{2}\ln\Bigg(\frac{\sigma_q^2+ \sigma_p^2+2k^2}{2\sqrt{\sigma_p^2+k^2}\sqrt{\sigma_q^2+k^2}}\Bigg) 
            \text{~.}
        \end{aligned}
    \end{equation*}
    As noise increases, $k$ becomes larger, and so:
    \begin{equation*}
        \begin{aligned}
            \lim_{k\to\infty} \frac{(\mu_p-\mu_q)^2}{\sigma_q^2+\sigma_p^2+2k^2} &= 0 \\
            \lim_{k\to\infty} \frac{\sigma_q^2+ \sigma_p^2+2k^2}{2\sqrt{\sigma_p^2+k^2}\sqrt{\sigma_q^2+k^2}} &= 1
            \text{~,}
        \end{aligned}
    \end{equation*}
    which gives:
    \begin{equation*}
        \lim_{k\to\infty} D_B(p+\delta_k,q+\delta_k) = 0
            \text{~.}
    \end{equation*}
\end{proof}

\subsection{Robustness Ratio\label{apx:ratio}}

If $\boldsymbol{x}_i$ is a sample from a data set $\mathbf{X}\subseteq\mathbb{R}^{N\times D}$, we can approximate $R_k(f, M, \mathbf{X})$ using the discrete formulation:
\begin{equation*}
    R_k^{[K]}(f,M,\mathbf{X}) =\sum_{j=1}^K\sum_{i=1}^N\frac{M(f,\tilde{\boldsymbol{x}}_{i,j})}{M(f,\boldsymbol{x_i})}
            \text{~.}
\end{equation*}

\begin{proof}
Take:
\begin{equation*}
    R_k^{[K]}(f,M,\mathbf{X}) = 1 + \frac{1}{KN} \sum_{j=1}^K\sum_{i=1}^N\frac{M(f,\tilde{\boldsymbol{x}}_{i,j})-M(f,\boldsymbol{x_i})}{M(f,\boldsymbol{x_i})}
            \text{~,}
\end{equation*}
where $\tilde{\boldsymbol{x}}_{i,j}$ is the $j$\textsuperscript{th} perturbation (with a strength $k$) of $\boldsymbol{x}_i$. Due to the law of large numbers:
\begin{equation*}
    R_k^{[K]}(f, M, \mathbf{X}) \to R_k(f, M, \mathbf{X}) \quad \textrm{as} \quad K\to\infty
            \text{~.}
\end{equation*}
Therefore:
\begin{equation*}
\begin{aligned}
    R_k^{[K]}(f,M,\mathbf{X}) &= 1 + \frac{1}{KN} \sum_{j=1}^K\sum_{i=1}^N\frac{M(f,\tilde{\boldsymbol{x}}_{i,j})-M(f,\boldsymbol{x_i})}{M(f,\boldsymbol{x_i})} \\
    &= 1 + \frac{1}{KN} \bigg( \sum_{j=1}^K\sum_{i=1}^N\frac{M(f,\tilde{\boldsymbol{x}}_{i,j})}{M(f,\boldsymbol{x_i})} - 1\bigg) \\
    &=1 + \frac{1}{KN} \bigg( \sum_{j=1}^K\sum_{i=1}^N\frac{M(f,\tilde{\boldsymbol{x}}_{i,j})}{M(f,\boldsymbol{x_i})}\bigg) - \frac{KN}{KN} \\
    &=\frac{1}{KN} \sum_{j=1}^K\sum_{i=1}^N\frac{M(f,\tilde{\boldsymbol{x}}_{i,j})}{M(f,\boldsymbol{x_i})}
            \text{~.}
\end{aligned}
\end{equation*}
\end{proof}

\subsection{Bias Bound under Equalised Odds\label{apx:bias_bound}}

\begin{proof}
Take
\begin{equation*}
    \beta_{y,a}=\mathbb{P}\{\mathbf{A}=1\vert\mathbf{Y}=y,\mathbf{A}_c=a\}
            \text{~.}
\end{equation*}
From \citet{eotheory} we already know that
\begin{equation}
    \begin{aligned}
        M_{eo}(\mathbf{F},\mathbf{X})\big\vert_{\mathbf{Y}=1}=M_{eo}(f,\mathbf{X})\big\vert_{\mathbf{Y}=1}\vert\beta_{1,0}(p_{1,0}-p_{0,0})+\beta_{1,1}(p_{1,1}-p_{0,1})\vert
    \end{aligned}
    \label{eou}
\end{equation}
and
\begin{equation*}
    \begin{aligned}
        M_{eo}(\mathbf{F},\mathbf{X})\big\vert_{\mathbf{Y}=0}=M_{eo}(f,\mathbf{X})\big\vert_{\mathbf{Y}=0}\vert\beta_{0,0}(p_{1,0}-p_{0,0})+\beta_{0,1}(p_{1,1}-p_{0,1})\vert
            \text{~.}
    \end{aligned}
\end{equation*}
Using Bayes theorem~\cite{bayes}, we can therefore see that
\begin{equation}
    \begin{aligned}
        \beta_{y,a}&=\mathbb{P}\{\mathbf{A}=1\vert\mathbf{Y}=y,\mathbf{A}_c=a\} \\
        &=\frac{\mathbb{P}\{\mathbf{A}=1, \mathbf{Y}=y\vert\mathbf{A}_c=a_c\}}{\mathbb{P}\{\mathbf{Y}=y\vert\mathbf{A}_c=a_c\}}\\
        &=\frac{\mathbb{P}\{\mathbf{A}=1,\mathbf{Y}=y,\mathbf{A}_c=a\}}{\mathbb{P}\{\mathbf{Y}=y,\mathbf{A}_c=a\}} \\
        &=\frac{1-k}{\mathbb{P}\{\mathbf{Y}=y,\mathbf{A}_c=a\}}
            \text{~.}
    \end{aligned}
    \label{bayeseq}
\end{equation}
From Equations~\ref{eou} and \ref{bayeseq} we can see that
\begin{equation}
\begin{aligned}
    M_{eo}(\mathbf{F},\mathbf{X})\big\vert_{\mathbf{Y}=1}=&M_{eo}(f,\mathbf{X})\big\vert_{\mathbf{Y}=1}(1-k)\cdot \\ &\Bigg\vert\frac{p_{1,0}-p_{0,0}}{\mathbb{P}\{\mathbf{Y}=1,\mathbf{A}_c=0\}}+\frac{p_{1,1}-p_{0,1}}{\mathbb{P}\{\mathbf{Y}=1,\mathbf{A}_c=1\}}\Bigg\vert \\
    =&\alpha M_{eo}(f,\mathbf{X})\big\vert_{\mathbf{Y}=1}(1-k)
            \text{~.}
    \label{boundproof}
    \end{aligned}
\end{equation}
Therefore, if $k=0$ (i.e., no noise) we get the upper bound from Equation~\ref{bounds}, so
\begin{equation*}
    M_{eo}(\mathbf{F},\mathbf{X})\big\vert_{\mathbf{Y}=1} \leq \alpha M_{eo}(f,\mathbf{X})\big\vert_{\mathbf{Y}=1}
            \text{~.}
\end{equation*}
We also know that $\mathbb{P}\{\mathbf{Y}=1,\mathbf{A}_c=a\}\in[0,1]$, and so
\begin{equation}
\frac{a}{\mathbb{P}\{\mathbf{Y}=1,\mathbf{A}_c=a\}} \geq a
            \text{~,}
\label{geqeq}
\end{equation}
giving
\begin{equation*}
    \begin{aligned}
        M_{eo}(\mathbf{F},\mathbf{X})\big\vert_{\mathbf{Y}=1}\geq M_{eo}(f,\mathbf{X})\big\vert_{\mathbf{Y}=1}(1-k)(p_{1,0}-p_{0,0}+p_{1,1}-p_{0,1})
            \text{~,}
    \end{aligned}
\end{equation*}
which is the lower bound from Equation~\ref{bounds}.
\end{proof}

\begin{lemma}
    Under the same conditions, as the noise $k$ increases the lower bound decreases.
\end{lemma} 

\begin{proof}
It is clear that as $k\to1$, $(k-1)\to0$, and therefore
\begin{equation*}
    (1-k)(p_{1,0}-p_{0,0}+p_{1,1}-p_{0,1})\to0
            \text{~.}
\end{equation*}
\end{proof}

\begin{lemma}
    Under the same conditions, as the noise $k$ increases $M_{eo}(f,\mathbf{X})\big\vert_{\mathbf{Y}=1}$ decreases, but will not decrease beyond $M_{eo}(f,\mathbf{X})\big\vert_{\mathbf{Y}=1}(1-k)(p_{1,0}-p_{0,0} + p_{1,1}-p_{0,1}$).
\end{lemma}

\begin{proof}
    Following from Equations~\ref{boundproof} and \ref{geqeq}, we can see that 
    \begin{equation*}
        \big\vert \mathbb{P}\{\mathbf{Y}=1,\mathbf{A}_c=0\} - \mathbb{P}\{\mathbf{Y}=1,\mathbf{A}_c=1\}\big\vert \to 0
            \text{~.}
    \end{equation*}
    This means that the denominator in Equation~\ref{boundproof} become equal, giving
    \begin{equation*}
        \begin{aligned}
            M_{eo}(\mathbf{F},\mathbf{X})\big\vert_{\mathbf{Y}=1}=M_{eo}(f,\mathbf{X})\big\vert_{\mathbf{Y}=1}(1-k)\cdot\Bigg\vert\frac{p_{1,0}-p_{0,0}+p_{1,1}-p_{0,1}}{\mathbb{P}\{\mathbf{Y}=1,\mathbf{A}_c=0\}}\Bigg\vert
            \text{~.}
        \end{aligned}
    \end{equation*}
    From Equation~\ref{geqeq}, we can therefore see that the smallest value is
    \begin{equation*}
        M_{eo}(f,\mathbf{X})\big\vert_{\mathbf{Y}=1}(1-k)(p_{1,0}-p_{0,0} + p_{1,1}-p_{0,1})
            \text{~.}
    \end{equation*}
\end{proof}

\subsection{Maximum Demographic Parity Due to Noise\label{apx:max}}

Given a model $f:\mathcal{X}\mapsto\mathcal{Y}$ and a threshold optimiser $F:f(\mathcal{X})\mapsto\mathcal{Y}$, the unfairness limit due to noise is given by:
    \begin{equation*}
    \begin{aligned}
        M_{dp}(F,\mathbf{\tilde{X}}_\infty) = \big\vert&(p_{1,0}\mathbb{P}\{f(\mathbf{\tilde{X}}_\infty)\geq T_{1,0}\} + p_{0,0}\mathbb{P}\{f(\mathbf{\tilde{X}}_\infty)\geq T_{0,0}\}) \\- &(p_{0,1}\mathbb{P}\{f(\mathbf{\tilde{X}}_\infty)\geq T_{0,1}\} + p_{1,1}\mathbb{P}\{f(\mathbf{\tilde{X}}_\infty)\geq T_{1,1}\})\big\vert
            \text{~.}
        \end{aligned}
    \end{equation*}

\begin{proof}
If $f:\mathbf{X}\mapsto [b_1,b_2]$, then, as $\mathbf{X}$ becomes increasingly random, $f(\mathbf{\tilde{X}_\infty})\sim U([b_1,b_2])$. If the output of $f$ is now uniformly random, then:
\begin{equation*}
    \begin{aligned}
        \mathbb{P}\{f(\mathbf{\tilde{X}}_\infty) \geq T_{y,a}\vert\mathbf{A}=a\}=
        \begin{cases}
        \frac{b_2-T_{y,a}}{b_2-b_1} &\textrm{ if } T_{y,a} \in [b_1, b_2] \\
        1 &\textrm{ if } T_{y,a} < b_1 \\
        0 &\textrm{ if } T_{y,a} > b_2
            \text{~.}
        \end{cases}
    \end{aligned}
\end{equation*}
Therefore, from Equation~\ref{eq:probs}:
\begin{equation*}
        \begin{aligned}
            \mathbb{P}\{F(\mathbf{\tilde{X}}_\infty) = 1\vert\mathbf{A} = a\} = \hspace{0.07cm}&p_{1,a}\mathbb{P}\{f(\mathbf{\tilde{X}}_\infty) \geq T_{1,a}\vert\mathbf{A}=a\} \\+\hspace{0.07cm}  &p_{0,a}\mathbb{P}\{f(\mathbf{\tilde{X}}_\infty) \geq T_{0,a}\vert\mathbf{A}=a\}
            \text{~,}
        \end{aligned}
    \end{equation*}
    which gives Equation~\ref{maxrand} when substituted into Definition~\ref{dp}. 
\end{proof}





\end{document}